\documentclass[12pt,A4]{article}
\usepackage[margin=1in,footskip=0.25in]{geometry}
\usepackage{subfig}
\usepackage{graphicx, color}  

\usepackage{algorithm}
\usepackage{algpseudocode}

\usepackage{bm}
\usepackage{amsmath, amsfonts}

\usepackage[comma]{natbib}
\usepackage{url}
\usepackage{enumerate}

\newtheorem{theorem}{Theorem}

\newtheorem{lemma}{Lemma}

\newcommand{\tabincell}[2]{\begin{tabular}{@{}#1@{}}#2\end{tabular}}  

\def\ni{\noindent}

\def\x{\bm{x}}   
 \def\w{\bm{w}}  \def\v{\bm{v}} 

\def\W{\bm{W}} \def\I{\bm{I}}  \def\A{\bm{A}}
 
\def\G{\bm{G}} \def\H{\bm{H}}

 \def \bb{\bm\beta} 
 \def\thth{\bm\theta}

\def \LL {\mathcal{L}}

\def \RRR {\mathbb{R}} \def \EEE {\mathbb{E}}

\begin{document}
	
	\title{\bf Enhancing Explainability of Neural Networks through Architecture Constraints}
	
	\author{Zebin Yang$^1$, Aijun Zhang$^1$ and Agus Sudjianto$^2$\\
		{\normalsize  $^1$Department of Statistics and Actuarial Science, The University of Hong Kong,}\\
		{\normalsize Pokfulam Road, Hong Kong}\\
		{\normalsize $^2$Corporate Model Risk, Wells Fargo, USA}}
	\date{}
	
	\maketitle
	
	\begin{abstract}
		Prediction accuracy and model explainability are the two most important objectives when developing machine learning algorithms to solve real-world problems. The neural networks are known to possess good prediction performance, but lack of sufficient model interpretability. In this paper, we propose to enhance the explainability of neural networks through the following architecture constraints:  a) sparse additive subnetworks; b)  projection pursuit with orthogonality constraint; and c) smooth function approximation. It leads to an explainable neural network (xNN) with the superior balance between prediction performance and model interpretability.	We derive the necessary and sufficient identifiability conditions for the proposed xNN model. The multiple parameters are simultaneously estimated by a modified mini-batch gradient descent method based on the backpropagation algorithm for calculating the derivatives and the Cayley transform for preserving the projection orthogonality. Through simulation study under six different scenarios, we compare the proposed method to several benchmarks including least absolute shrinkage and selection operator, support vector machine, random forest, extreme learning machine, and multi-layer perceptron. It is shown that the proposed xNN model keeps the flexibility of pursuing high prediction accuracy while attaining improved interpretability. Finally, a real data example is employed as a showcase application.
		\vskip 6.5pt \noindent {\bf Keywords}: 
		Explainable neural network, additive decomposition, sparsity, orthogonality, smoothness.
	\end{abstract}
	
	\section{Introduction}
	The recent developments of neural network techniques offer tremendous breakthroughs in machine learning and artificial intelligence (AI). The complex network structures are designed and have brought great successes in areas like computer vision and natural language processing. Besides predictive performance, transparency and interpretability are essential aspects of a trustful model; however, most of the neural networks remain black-box models, where the inner decision-making processes cannot be easily understood by human beings. Without sufficient interpretability, their applications in specialized domain areas such as medicine and finance can be largely limited.  For instance, a personal credit scoring model in the banking industry should be not only accurate but also convincing. The terminology ``Explainable AI'' advocated by the Defense Advanced Research Projects Agency (DARPA) draws public attention \citep{gunning2017explainable}. Recently, the US Federal Reserve governor raised the regulatory use of AI in financial services \citep{Brainard2018} and emphasized the development of explainable AI tools.

	There has been a considerable amount of research works on interpretable machine learning by post-hoc analysis, including the model-agnostic approach and the model distillation approach. The examples of the former approach are the partial dependence plot \citep{friedman2001greedy}, the individual conditional expectation plot \citep{goldstein2015peeking}, the locally interpretable model-agnostic explanation method \citep{ribeiro2016should} and others \citep{apley2016visualizing, liu2018model}. The examples of the latter approach are model compression and distillation \citep{bucilua2006model, ba2014deep}, network distilling \citep{hinton2015distilling}, network pruning \citep{wang2018novel}, and tree-partitioned surrogate modeling \citep{hu2018locally}. 

	It is our goal to design highly predictive models that are intrinsically interpretable. This is a challenging task as the two objectives (i.e., prediction accuracy and model 	interpretability) usually conflict with each other \citep{gunning2017explainable}. A deep neural network may provide accurate prediction, but it can be hardly understood even by the model developers. In contrast, most statistical models are intrinsically explainable (e.g., linear regression, logistic regression, and decision tree), but these simplified models are known to be less predictive than the black-box type of machine learning algorithms in dealing with large-scale complex data. We find that the additive index model (AIM) is a promising statistical model for balancing prediction accuracy and model interpretability. 

	The AIM, also known as the projection pursuit regression (PPR; \citealp{friedman1981projection}), decomposes a complex function into the linear combination of multiple component functions. Given a dataset $\{\x_{i},y_{i}\}_{i\in[n]}$ with features $\bm{x} \in \mathbb{R}^{p}$ and response $y$, the AIM takes the following form, 
	\begin{equation} \label{AIM}
	g(\EEE(y|\x)) = \mu + h_{1}(\bm{w}_{1}^{T}\bm{x})  + \ldots + h_{k}(\bm{w}_{k}^{T}\bm{x}),
	\end{equation}
	where $g$ is a link function, $\mu$ is the intercept, $\w_j\in\RRR^p$ are the projection indexes for $j=1,\ldots,k$ and $h_j(\bm{w}_{j}^{T}\bm{x})$ are the corresponding nonparametric ridge functions. Note that the AIM is closely related to neural networks~\citep{hwang1994regression}. If we fix each ridge function to be a prespecified activation function, it reduces to a single-hidden-layer neural network. Indeed, the AIM is also a universal approximator as $k$ is sufficiently large. 

	The estimation of AIM is usually based on alternating optimization between the ridge functions and the projection indexes. When $\W = [\w_{1},\ldots, \w_{k}]$ is fixed, the ridge functions are estimated by nonparametric smoothers subject to backfitting. When $\{h_j\}_{j\in [k]}$ are fixed, the projection indexes are iteratively estimated by Gauss-Newton algorithms. Such an alternating procedure may not guarantee the global optimum, and it becomes extremely time-consuming for large-sample datasets. To improve computation, an AIM-based neural network was recently proposed by~\cite{vaughan2018explainable}, where each activation function of the hidden layer is extended to be a ridge function $h_j(\cdot)$ as modeled by a feedforward subnetwork. Each subnetwork consists of one input node, multiple hidden layers, and one output node. With such a network architecture, the AIM parameters can be simultaneously estimated by mini-batch gradient descent, which is effective and scalable for large-scale datasets. 

	In~\cite{vaughan2018explainable}, the above AIM-based neural network is called an {\em explainable} neural network (xNN), which is essentially an AIM with NN-represented ridge functions. The additive model structure of (\ref{AIM}) makes it easy to explain the effect attribution only when the ridge functions are mutually independent. However, such an independence condition is hard to be satisfied, especially when $k$ grows. Actually, as discussed in Chapter~11 of~\cite{hastie2009elements}, the AIM is usually regarded as a non-interpretable model. We argue that the model interpretability should be induced from practical constraints, and an AIM is not explainable unless it meets further interpretability constraints. In this regard, the xNN by~\cite{vaughan2018explainable} can achieve partial explainability as it is suggested to be equipped with $\ell_1$ shrinkage for the purpose of sparse and parsimonious modeling. We call it a naive version of explainable neural network (xNN.naive).

	There are some potential drawbacks of the naive xNN. First, it was suggested to rewrite each ridge function as $\beta_j \tilde{h}_j(\w^T\x)$ and then apply the $\ell_1$ shrinkage on $\bb = \{\beta_j\}_{j\in[k]}$ to enforce the sparsity. Such split ridge functions become unidentifiable unless appropriate norm constraints are imposed. Second, the projection weights $\w_1,\ldots,\w_k$ in the original AIM  can be correlated, hence violate the independence condition for clear effect attribution. It is desirable that the resulting projection indexes are mutually orthogonal. Third, the ridge functions represented by the neural networks may not be as smooth as the nonparametric methods (e.g., the smoothing splines). The non-smooth ridge functions would also add difficulty for explaining the functional relationship. 

	In this paper, we propose to enhance the explainability of neural networks through architecture constraints, including additivity, sparsity, orthogonality, and smoothness. Specifically, the classical fully connected network architecture is re-configured with sparse additive subnetworks subject to $\ell_1$-sparsity constraints on both the projection layer and the output layer, together with an additional normalization step for ensuring ridge function identifiability. The orthogonality constraint is imposed on the projection indexes such that $\W^T\W = \I_k$, which is known as the Stiefel manifold $\mbox{St}(p,k)$ and can be treated by the Cayley transform~\citep{wen2013feasible}. For each ridge function, the smoothness constraint is imposed by considering an empirical roughness penalty based on the squared second-order derivatives~\citep{wahba1990spline}. Combining these interpretability constraints into the neural network architecture, we obtain an enhanced version of explainable neural network (xNN.enhance).

	Computationally, the enhanced xNN model is estimated by modern neural network training techniques, including backpropagation, mini-batch gradient descent, batch normalization, and the Adam optimizer. We develop a new SOS-BP algorithm based on the backpropagation technique for calculating the derivatives and the Cayley transform for preserving the projection orthogonality. All the unknown parameters in the enhanced xNN model can be simultaneously optimized by mini-batch gradient descent. Owing to the modern machine learning platform (e.g., the \textsl{TensorFlow} used here) with the automatic differentiation technology, the empirical roughness penalty can be readily evaluated for each projected data point. The involved hyperparameters can be either set to our suggested configurations or optimized by grid search, random search, or other automatic procedures. Moreover, the proposed SOS-BP algorithm based on the mini-batch training strategy is scalable to large-scale datasets, and it can also be implemented on GPU machines. 	

	The enhanced xNN architecture keeps the flexibility of pursuing prediction accuracy while attaining the improved interpretability. It can be therefore used as a promising surrogate for complex model approximation. Through simulation study under six different scenarios, its prediction accuracy is shown to outperform or perform quite close to classical machine learning models, e.g., support vector machine (SVM) and multi-layer perceptron (MLP). On the other hand, the enhanced xNN model conveys the intrinsic interpretability in terms of projection weights and corresponding subnetworks. As a showcase of potential applications, the proposed model is used to study a real data example from LendingClub. 

	This paper is organized as follows. Section~\ref{Methodology} provides the AIM-based formulation of the explainable neural network subject to sparsity, orthogonality, and smoothness constraints. The identifiability conditions for the proposed xNN model are derived. Section~\ref{Estimation} discusses the computational method through the new SOS-BP algorithm. Numerical experiments are conducted with both simulation studies and a real data example, as presented in Section~\ref{Numerical}. Finally, Section~\ref{Conclusion} concludes the paper with remarks about future study.

	\section{Explainable Neural Networks} \label{Methodology}
	Given the feature vector $\x\in\RRR^p$ and the response $y\in \RRR$, consider the AIM with the matrix of projection indexes $\W = [\w_{1},\ldots, \w_{k}] \in \RRR^{p\times k}$ for $k\leq p$, and the corresponding normalized ridge functions $\tilde{h}_j(z)$ of each projected variable or learned feature $z=\w_j^T\x$, for $j=1,\ldots,k$. The proposed xNN model is formulated by:
	\begin{subequations}\label{SOSxNN}
		\begin{equation}\label{xNN}
		g(\EEE(y|\x)) =  \mu + \sum\limits_{j=1}^{k} \beta_j \tilde{h}_{j}(\w_j^T\x) + \varepsilon, \tag{\ref{SOSxNN}}
		\end{equation}
		subject to the following interpretability constraints:
		\begin{eqnarray}
		&  {\displaystyle \sum_{i=1}^p \big|W_{ij}\big| \leq T_1,} \label{ConA}\\
		& {\displaystyle \sum_{j=1}^k \big|\beta_j\big|  \leq T_2,} & \label{ConB}\\
		& {\displaystyle \int \big[\tilde{h}_{j}''(z)\big]^2dF_j(z)  \leq T_3,}  \label{ConC} \\
		&  \W^T\W = \I_k,  \label{ConD}\\
		& {\displaystyle \int \tilde{h}_j(z)dF_{j}(z)=0, \tilde{h}_j(z)^2dF_{j}(z)=1, \label{ConE}}
		\end{eqnarray}
	\end{subequations}
	for $j=1,\ldots, k$, where $T_1\geq 1$, $T_2, T_3 \geq 0 $ are the regularization strengths and each $F_{j}$ represents the empirical cumulative distribution function of the $j$th projected variable. The multiple constraints (\ref{ConA}-\ref{ConE}) are imposed by the interpretability considerations from the sparse, orthogonal, and smooth perspectives. The coefficients $\beta_{j}$ for $j=1,\ldots, k$ represent signed scales for each normalized ridge function, based on which we define the corresponding importance ratio (IR) as 
	\begin{equation}\label{ImRatio}
	r_{j} = |\beta_{j}| / \sum_{j=1}^{k} |\beta_{j}|, \quad  j=1,\ldots,k.
	\end{equation} 
	
\begin{figure*}[!t]
		\centering
		\includegraphics[width=0.72\textwidth]{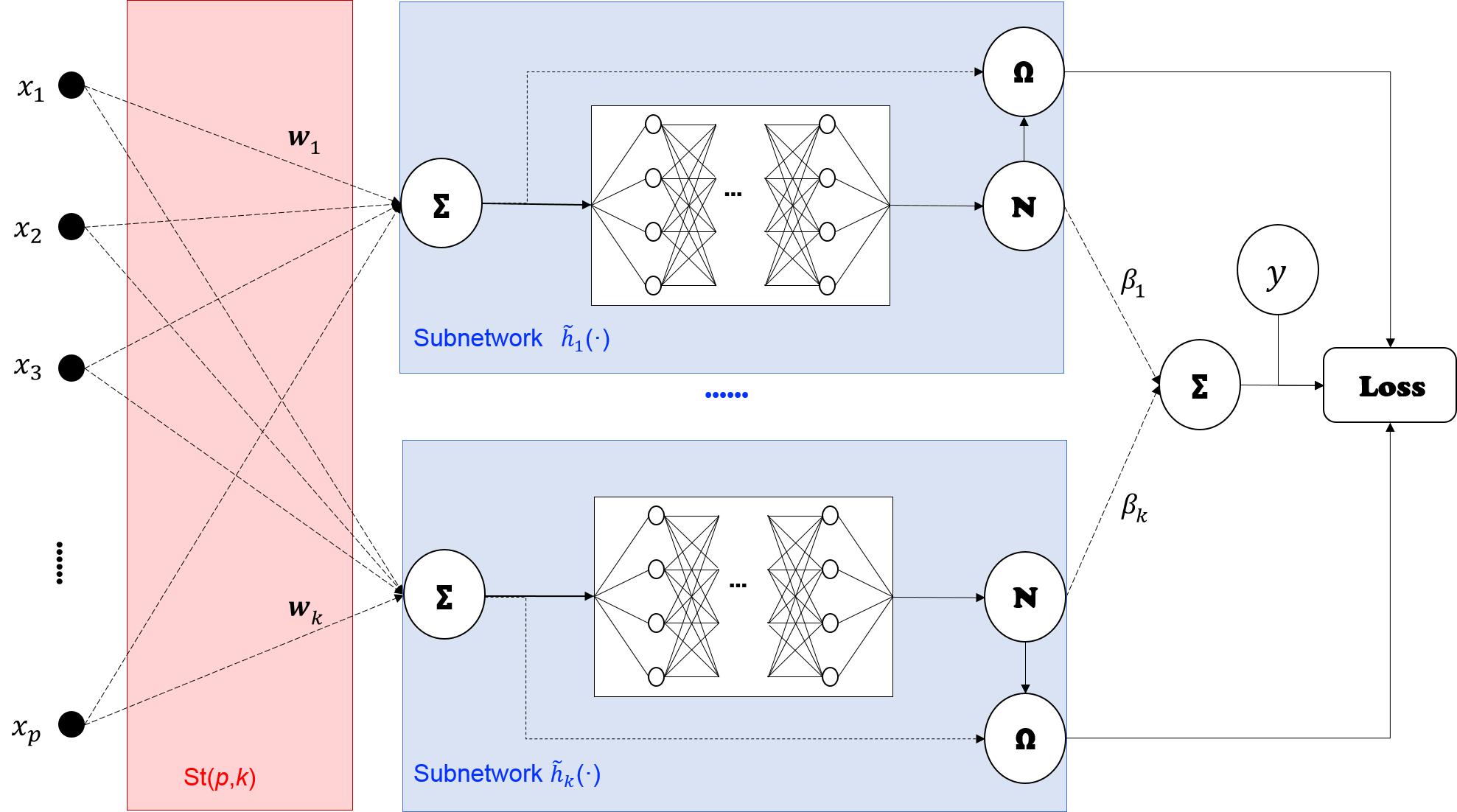}
		\caption{The enhanced xNN architecture: the dashed arrows to the $\bm\Sigma$ nodes are subject to the sparse constraints on $\W=\{\w_j\}_{j\in[k]}$ and $\bb=\{\beta_j\}_{j\in[k]}$, respectively; the red shaded $\mbox{St}(p,k)$ area represents the Stiefel manifold constraint on the projection indexes ($\W^T\W=\I_k$), and for each blue shaded subnetwork, a smooth function approximation is performed to capture the normalized ridge function through a normalization node ${\bf N}$ and a roughness penalty node $\bm\Omega$.}\label{fig:SOSxNN}
	\end{figure*}

	Fig.~\ref{fig:SOSxNN} presents the proposed xNN architecture using neural network notations. In particular, each node $\bm\Sigma$ represents the linear combination with inputs from the previous layer and the coefficient arrows, where the arrows are drawn as dashed lines to indicate that the coefficients are subject to sparsity constraints. The shaded area of $\mbox{St}(p,k)$ indicates the orthogonality constraint for the projection indexes $\W^T\W = \I_k$. For each subnetwork, it consists of one input node for projected data, multiple hidden layers for the feedforward neural network, one output node subject to ${\bf N}$ (normalization) and an additional node for calculating $\bm\Omega$ (roughness penalty). All the normalized ridge functions are then linearly combined to form the final output, together with a bias node for capturing the overall mean.  
	
	Table~\ref{notation_list} lists a summary of notations in both AIM and  xNN contexts, as well as the interpretation based on the enhanced xNN model. 
	
	\begin{table*}[t]
		\centering
		\small
		\renewcommand\tabcolsep{3pt}
		\renewcommand\arraystretch{1.0}
		\caption{Summary of notations in the AIM and the xNN contexts.} \label{notation_list}
		\begin{tabular}{c|c|c|c}
			\hline
			        Symbol         &            AIM            &                        xNN                        &                                 Interpretation based on xNN                                  \\ \hline
			        $\mu$          &         intercept         &            the bias of the output node            &                                         overall mean                                         \\ \hline
			       $\w_{j}$        &     projection index      &  \tabincell{c}{  weights of inputs \\ and the $j$th subnetwork}     &                \tabincell{c}{   coefficients for constructing \\the $j$th learned feature }                   \\ \hline
			$\tilde{h}_{j}(\cdot)$ & \tabincell{c}{normalized \\ ridge function} &               the $j$th subnetwork                & \tabincell{c}{the functional relationship of   the $j$th \\ learned feature and the response} \\ \hline
			      $\beta_j$        &   scaling	constant  & \tabincell{c}{ the weight of the $j$th \\ subnetwork and the output} &        \tabincell{c}{     the signed scale of \\ the $j$th normalized ridge function}        \\ \hline
		\end{tabular}
	\end{table*}

	\subsection{Interpretability Constraints} \label{Section_IdentCond}
	The explainability of the proposed xNN model is induced by the imposed interpretability constraints. We make justification from the following three aspects. 
	
	\textbf{a) Sparse Additive Subnetworks}
	
	The normalized ridge functions in (\ref{SOSxNN}) are modeled by the additive subnetworks, where each subnetwork corresponds to a feedforward neural network for flexible function approximation. A single-hidden-layer feedforward neural network with an arbitrarily wide hidden layer and appropriate activation functions is known to approximate a continuous function on a compact domain arbitrarily well~\citep{hornik1991approximation}. Therefore, it is flexible in specifying the number of layers and nodes for each subnetwork, according to the complexity of tasks.
	
	Two $\ell_1$-norm constraints are imposed for inducing the sparsity in ridge functions and projection weights, in order to make the model parsimonious. By the $\ell_1$ constraint (\ref{ConB}) upon a suitable choice of $T_2$, a certain number of ridge functions will have zero scales, i.e., $\hat\beta_j = 0$ for some $j$; such ridge functions and corresponding projection indexes will become inactive. By the $\ell_1$ constraint (\ref{ConA}) upon the suitable choice of $T_1$, the sparsity in each projection index can be induced such that some negligible weights will be regularized to zero, similar to the sparse principal component analysis by \cite{zou2006sparse}.
	
	For the sake of subnetwork identifiability, the overall mean and norm of functions are constrained by the condition (\ref{ConE}). This is essentially performing normalization of each estimated ridge function. It is a critical procedure for simultaneously estimating all the subnetworks while performing the $\ell_1$ shrinkage on the scales of ridge functions.

	\textbf{b) Orthogonal Projection Pursuit} 
	
	The constraint (\ref{ConD}) is imposed to enforce projection indexes mutually orthogonal, which is also known as the Stiefel manifold $\W\in\mbox{St}(p, k)$. It is a natural idea to consider mutually orthogonal projection indexes to avoid ambiguity due to multicollinearity, see, e.g., the principal component regression (PCR) by~\cite{jolliffe1982note} and the supervised PCA by~\cite{bair2006prediction, barshan2011supervised}. Without the orthogonality assumption, the AIM often results in correlated or even identical projection indexes, which makes it difficult to distinguish different additive components.  The orthogonality constraint can also be understood geometrically, as it provides an orthogonal basis for data rotation. Therefore, the proposed xNN model is more explainable than the AIM and the naive xNN model. 
	
	\textbf{c) Smooth Function Approximation} 
	
	The functional roughness penalty (\ref{ConC}) is imposed as a constraint in order to enforce smoothness of each ridge function. Following \cite{wahba1990spline}, we formulate it by the integrated squared second-order derivatives over the entire range of projected data $z=\w_j^T\x$. Using the empirical distribution, the integral form in (\ref{ConC}) can be rewritten as
	\begin{equation}\label{ConCa}
	\Omega(\tilde{h}_j) = \frac{1}{n}\sum_{i=1}^{n} \Big[\tilde{h}_j''(\w_j^T\x_i)\Big]^2, \mbox{ for } j=1,\ldots,k.
	\end{equation}
	Thus, the subnetworks subject to roughness penalty can be regarded as a neural network alternative  to nonparametric smoothers. As a benefit, the neural network training techniques can be directly used for fitting the smooth ridge functions. Compared with the naive xNN model without smoothness regularization, the enhanced xNN model can prevent non-smooth representation and achieve better interpretability.
	
	\subsection{Identifiability Conditions} \label{Identifiability_Conditions}
	A model is not identifiable if it has more than one representations. Such non-uniqueness will make a model hard to interpret. It turns out the interpretability constraints for the proposed xNN model can actually make the model more identifiable, which in turn enhances the model interpretability. In this section, we derive the identifiability  conditions for the enhanced xNN model.
	
	For the AIM, some trivial conditions should be imposed for removing possible ambiguity caused by shifting, scaling, and flipping effects, see \cite{yuan2011identifiability, chen2016generalized} for details. Then, we have the following lemma for the standard AIM. 
	
	\begin{lemma} \label{AIM_Condition}
		(Theorem 1 of \cite{yuan2011identifiability}). The AIM (\ref{AIM}) is identifiable if  the following conditions are satisfied:		
		\begin{itemize}		
			\item[] (A1) There is at most one linear ridge function. If $h_{j}$ is linear, then $\w_{j}^{T}\w_{i} = 0$ for all $j \neq i$;
			\item[] (A2) There is at most one quadratic ridge function;
			\item[] (A3) The projection matrix $\W$ is column full rank.
		\end{itemize}
	\end{lemma}
	
	After imposing the orthogonality constraint (\ref{ConD}), we prove that the enhanced xNN model has weaker identifiability conditions. Note that for $j=1,\ldots,k$ splitting of the ridge function $h_j(\cdot)$ into $\beta_j\tilde{h}_j(\cdot)$ does not affect model identifiability, since each $\tilde{h}(\cdot)$ is estimated subject to the normalization constraint (\ref{ConE}).  For convenience, we analyze the identifiability conditions based on the unnormalized form of ridge functions, as stated in the following theorem.  
	
	\begin{theorem} \label{THEOREM_Orthogonality} For the xNN model (\ref{SOSxNN}) with $\W^T\W = \I_k$,  
		it is identifiable if   the following conditions are satisfied:
		\begin{enumerate}
			\item[] (B1) There is at most one linear ridge function;
			\item[] (B2) If there exist multiple quadratic ridge functions, their quadratic term coefficients should be all distinct. 
		\end{enumerate}
	\end{theorem}
	
	Comparing Theorem~\ref{THEOREM_Orthogonality} to Lemma~\ref{AIM_Condition}, we still require at most one linear ridge function, since it is clear that any two linear ridge functions can be combined or represented with other two linear forms. Since we have imposed the orthogonality constraint in the enhanced xNN model, the rest part of Condition (A1) is naturally satisfied. Meanwhile, for Condition (A3), the projection matrix $\W$ is definitely column full rank. As for the requirement of the quadratic ridge functions in Condition (A2), it can be relaxed to Condition (B2). That is, multiple quadratic functions can be identified, as long as their quadratic term coefficients are all different.  The detailed proof of Theorem~\ref{THEOREM_Orthogonality} is available in  Appendix~\ref{proof_theorem1}.
	
	Theorem~\ref{THEOREM_Orthogonality} states the sufficient conditions for the xNN model to be identifiable. We have the following observations when examining the necessity of these conditions. First, without further assumption, the enhanced xNN model is not identifiable if there exist two or more linear ridge functions. For example, a linear term $ x_{1} $ can be expressed by two linear ridge functions $h_{1}(\w_{1}\x) + h_{2}(\w_{2}\x)$, where $h_{1}(z)=h_{2}(z)=\frac{\sqrt{2}}{2}z$, and $\w_{1} = [\frac{\sqrt{2}}{2}, \frac{\sqrt{2}}{2}]^{T}, \w_{2} = [\frac{\sqrt{2}}{2}, -\frac{\sqrt{2}}{2}]^{T}$. 
	
	Next, we consider the necessity of Condition (B2). Assume there are two quadratic functions $ h_{1}(z) = h_{2}(z) = \alpha z^{2} $ with mutually orthogonal projection indexes $ \W=[\w_{1},\w_{2}] $. The enhanced xNN model can be expressed in matrix notation as follows, 
	$$ h_{1}(\w_{1}\x) + h_{2}(\w_{2}\x) = \x^{T} \bm{U} \x,$$ 
	where $\bm{U}$ is a real symmetric matrix $\bm{U} = \alpha \W \W^{T}$. For any $k \times k$ matrix $\bm{V}$ subject to $\bm{V}\bm{V}^{T} = \bm{I}$ and $\bm{V} \neq \bm{I}$, we can construct a different projection matrix $\tilde{\W} = \W\bm{V}$ that
	$$
	\tilde{\bm{U}} = \alpha\tilde{\W}\tilde{\W}^{T}	= \alpha\W\bm{V}\bm{V}^{T}\W^T 
	= \alpha\W\W^{T}  
	= \bm{U}. 
	$$
	It is obvious that $\tilde{\W}^{T}\tilde{\W} = \bm{I}$ also holds, therefore the projection indexes are not identifiable.
	
	\subsection{Special Cases}
	There exist a few special cases of the enhanced xNNs which are worth mentioning below. 
	\begin{enumerate}
		\item If $T_1=1$, the xNN model reduces to the generalized additive model (GAM; \citealp{hastie1990generalized}); 
		\item If $T_1=1$ and $T_3=0$, the xNN model reduces to the generalized linear model (GLM);
		\item If $k=1$, the xNN model reduces to the single-index model (SIM; \citealp{ichimura1993semiparametric}).
	\end{enumerate}
	
	Given the orthogonality constraint (\ref{ConD}), it is easy to check that for each $j=1,\ldots,k$,  $\sum_{i=1}^{p} W_{ij}^2 =1$ and 
	$(\sum_{i=1}^{p} |W_{ij}|)^2 = \sum_{i=1}^{p} W_{ij}^2 + \sum_{i\neq i'} |W_{ij}| |W_{i'j}| \geq 1$. The equality $\sum_{i=1}^{p} |W_{ij}|=1$ holds only when there is exactly one non-zero element in the vector $[W_{1j},\ldots,W_{pj}]$. Therefore, when $T_1=1$ in the sparsity constraint (\ref{ConA}), each projection picks one component, and it corresponds to the GAM.  Furthermore, if $T_3=0$ in the smoothness constraint (\ref{ConC}), each ridge function is enforced to have zero second-order derivative, which reduces to the linear ridge functions as in the GLM. 
	
	The GAM, GLM, and SIM under these special cases are all identifiable and highly explainable. Furthermore, for the GAM (when $T_1=1$) subject to the sparsity constraint (\ref{ConB}), we obtain the sparse GAM~\citep{ravikumar2009sparse}. For the SIM (when $k=1$) subject to the sparsity constraint (\ref{ConA}), we obtain the SIM with sparse projection. These sparse models tend to be more parsimonious and even more explainable, and they are especially useful for high-dimensional  problems.

	\section{Computational Method} \label{Estimation}
	In this section, we discuss the computational procedures for estimating the enhanced xNN model (\ref{SOSxNN}). Denote the list of unknowns in the proposed model by 
	\begin{equation}\label{thth}
	\thth=\Big\{\mu; \beta_1, \ldots,\beta_k; \tilde{h}_1, \ldots, \tilde{h}_k; \w_1,\ldots,\w_k\Big\},
	\end{equation}
	consisting of scalars, vectors, and functions. It is noted that the normalized ridge function is approximated by the feedforward neural network with finite parameters, but for simplicity, we denote the ridge function as unknown. For each feature vector $\x$, denote the xNN prediction by $\hat{y} = \EEE(y|\x;\thth)$. To measure the prediction accuracy, a convex loss $l(\thth)$ can be specified depending on the type of response $y$, e.g., the least squares loss for regression and the cross-entropy loss for binary classification.
	
	By the method of Lagrange multipliers, both the $\ell_1$ penalty (\ref{ConA}, \ref{ConB}) and the $\ell_2$ roughness penalty (\ref{ConCa}) can be formulated as soft regularizers, while the orthogonality (\ref{ConD}), zero-mean and unit-norm requirements (\ref{ConC}) are hard constraints. It leads to the following constrained optimization problem, 
	\begin{equation} \label{OptProb}
		\begin{gathered}
		\min_{\thth}\LL_{\lambda_{1}, \lambda_{2}, \lambda_{3}}(\thth)  =  l(\thth) + \lambda_{1}\sum_{j=1}^{k}\|\bm{w}_{j}\|_{\ell_1} + \\
		\lambda_{2}  \|\bb\|_{\ell_1} + \lambda_{3} \sum_{j=1}^{k} \Omega(\tilde{h}_j) \\
		\mbox{s.t. }   \bm{W}^{T}\bm{W}=\I_k, \\ 
		{\displaystyle  \int \tilde{h}_j(z)dF_{j}(z)=0 }, {\displaystyle \int \tilde{h}_j(z)^2dF_{j}(z)=1},
		\end{gathered}
	\end{equation} 
	for $j=1,\ldots, k$, where $\lambda_{1}, \lambda_{2}, \lambda_{3} \geq 0$ are the regularization hyperparameters corresponding to the soft constraints (\ref{ConA}--\ref{ConC}), respectively.
	
	\subsection{SOS-BP Algorithm}
	To deal with the hard constraints in (\ref{OptProb}), we employ the Cayley transform for preserving projection indexes on the Stiefel manifold $ \bm{W}^{T}\bm{W}=\bm{I}_k$ and employ batch normalization for each estimated ridge function. The multiple unknown parameters in (\ref{thth}) are simultaneously optimized by mini-batch gradient descent, and such a backpropagation procedure is applied throughout the remaining discussions. 
	
	During the updates, the orthogonality of matrix $\W$ should be preserved, for which we employ a fast and effective update scheme proposed by \cite{wen2013feasible}. It is based on the following Cayley transform along gradient descent direction on the Stiefel manifold
	\begin{equation}\label{Cayley}
	\W(\tau)=\left(\bm{I}+\frac{\tau}{2}\bm{A} \right)^{-1}\left(\bm{I}-\frac{\tau}{2}\A \right)\W,
	\end{equation}
	where $\tau$ is a step size, $\A = \G_{\W}\bm{W}^{T} - \bm{W}\G_{\W}^{T}$ is a skew-symmetric matrix with $\G_{\W}$ being the partial gradient  $\partial \LL / \partial \W$. It can be verified that $\W(\tau)^T\W(\tau)=\W^T\W$ for $\tau\in\RRR$, and the orthogonality can be preserved as long as $ \bm{W}^{T}\bm{W}=\bm{I}_k$. It can also be justified that when the step size $\tau$ is positive and small enough, the objective function will get improved after the Cayley transform. 
	
	The multiple parameters in (\ref{thth}) can be simultaneously optimized by mini-batch gradient descent together with the Cayley transform. It is an iterative procedure with flexibility in specifying the mini-batch size $n_b$ and the number of epochs $M$. The total number of gradient descent iterations is controlled by $M$ multiplied by $\lfloor n/n_b \rfloor$. Within each iteration, the weight matrix $\W = \{\w_1,\ldots,\w_k\}$ can be updated separately by the Cayley transform, while the remaining parameters $ \thth\setminus\W$ (all the parameters except for $\W$) are updated by gradient descent with adaptive learning rates determined by the Adam optimizer~\citep{kingma2014adam}. The new SOS-BP algorithm for estimating the xNN model is presented in Algorithm~\ref{algo:SOS-BP}.

	\begin{algorithm}[h]
		\caption{The SOS-BP Algorithm} \label{algo:SOS-BP}
		\begin{algorithmic}
			\Require $\{(\x_i, y_i\}_{i\in[n]}$ (Training data), $k$ (Number of  subnetworks),  
			$\lambda_{1}, \lambda_{2}$ (Sparsity), $\lambda_{3}$ (Smoothness),
			$\H$ (Subnetwork structure), $\eta$ (Learning rate), $\tau$ (Step size for the Cayley transform),  
			$ n_{b} $ (Mini-batch size) and $M$ (Number of epochs). 
			\State Initialize the network with $\bm{W}$ satisfying $\bm{W}^{T}\bm{W}=\I_k$.
			\For{{\rm Epoch} $m = 1, ..., M$}  
			\State	Split the reshuffled  data into $B = \lfloor \frac{n}{n_{b}} \rfloor$ mini-batches, each with $n_{b}$ samples.
			\For{{\rm Batch} $b = 1, ..., B$}
			\State Select the $b$th mini-batch and set $t=(m-1)B + b$.			
			\State Update $\W$ by $\W^{(t+1)} = \W^{(t)}(\tau)$.
			\State Update $\bm{\tilde\theta}^{(t+1)}=\bm{\tilde\theta}^{(t)} - \eta_t\cdot \nabla_{\bm{\tilde\theta}}^{(t)}$, where $\bm{\tilde\theta} = \thth\setminus\W$.
			\State Perform batch normalization for $\tilde{h}_j$, $j=1,\ldots,k$.
			\State Adjust $\eta_t$  by the Adam optimizer.
			\EndFor
			\State  Stop if the validation score is not improving. 
			\EndFor
		\end{algorithmic}
	\end{algorithm}
	
	The proposed SOS-BP algorithm adopts the mini-batch gradient descent strategy, and it utilizes some of the latest developments of neural network training techniques. It is capable of handling very big dataset. Next, we discuss several other computational aspects. 
	\begin{enumerate}[a)]
		\item For initialization, the projection matrix $\W$ should be generated subject to the orthogonality constraint $\W^{T}\W=\I_k$, which can be achieved by the QR decomposition.
		
		\item Each subnetwork modeled by the feedforward neural network can be  parametrized by $\H = [n_1, n_2, \ldots; \mbox{``act-type''}]$, where $n_j$ stands for the number of nodes for the $j$th hidden layer and ``act-type'' is the type of activation function in each subnetwork. A deeper network could be more expressive but is also more expensive for training.
		
		\item The roughness penalty (\ref{ConCa}) for each ridge function is evaluated empirically for each mini-batch data, where the second-order derivatives for each data point are readily obtainable by automatic differentiation. This procedure corresponds to the dashed input arrow of the $\bm\Omega$ node within each subnetwork; see Fig.~\ref{fig:SOSxNN}.
		
		\item The three soft regularization terms in (\ref{OptProb}) are automatically taken into account by the Cayley transform and gradient descent. More specifically, $\sum_{j=1}^k\|\w_j\|_{\ell_1}$ takes effect when computing the partial gradient $\G_{\W}$ for the Cayley transform, while $ \|\bb\|_{\ell_1} $ and $\sum_{j=1}^{k} \Omega(\tilde{h}_j)$ take effect when computing the partial gradient $ \nabla_{\bm{\tilde\theta}}^{(t)} = \partial \LL / \partial \bm{\tilde\theta}$ in Algorithm~\ref{algo:SOS-BP}.
		
		\item To deal with the zero-mean and unit-norm constraint for each ridge function, a normalization procedure is required. We adopt the popular batch normalization strategy with momentum set to zero. Referring to Fig.~\ref{fig:SOSxNN}, the batch normalization is performed by the {\bf N} node within each subnetwork. 
	\end{enumerate}
	
	The $\ell_1$ regularization in neural networks is based on the weight decay technique, which is used in our SOS-BP algorithm for shrinking the number of ridge functions or subnetworks. Unlike the $\ell_1$ shrinkage in the sparse linear modeling (e.g., LASSO), the weight decay method cannot exactly enforce the weights of small effects to zero. Therefore, a subnetwork pruning procedure is needed to filter out the subnetworks with sufficiently small $|\beta_{j}|,~j=1,2,\ldots,k$. As a rule of thumb, we can rank all the subnetworks according to their importance ratios (\ref{ImRatio}), and select the top-ranked subnetworks that accumulate more than 95\% or 99\% of importance ratios. The rest subnetworks are treated as negligible, and their corresponding $\beta_{j}$ are set to zero. We also apply a one-step refining procedure to re-estimate the selected subnetworks, conducted as follows, a) fix the projection layer; b) fix the regularization strength $\lambda_{1} = \lambda_{2} = 0$; c) fine-tune the remaining network via the Adam optimizer for certain epochs. After fine-tuning finishes, we can recalculate the importance ratios, and visualize the estimated projection indexes and ridge functions/subnetworks.	
	
	\subsection{Hyperparameter Settings} \label{Estimation:hyperpara}
	As in typical neural network models, there exist multiple hyperparameters that need to be configured. In this section, we provide a detailed guideline for the hyperparameter configuration for the proposed xNN model.
			
	For parsimonious modeling purpose, a small number of subnetworks is preferable. For instance, we can set $ k = \min\{p, 10\} $ in practice. For the architecture of subnetworks, a network with two hidden layers and nonlinear activations is tested to be sufficient for curve fitting, and we fix the subnetwork structure to $ [10, 6; ``tanh"] $ with hyperbolic tangent activations. 
	
	For gradient descent, the projection weights are initialized to be an orthogonal matrix, and the weights of the rest layers are initialized with Xavier normal initializer~\citep{glorot2010understanding}. The initial learning rate $\eta$ is tested to be not sensitive to the performance as it is sufficiently small, and we fix it to 0.001. It is worth mentioning that for the Cayley transform, an adaptive scheme for the step size $\tau$ was proposed by \cite{wen2013feasible}, while here, we fix $\tau=0.1$ to be a relatively small value, which makes the network training relatively straightforward. For training data with sample size $n$, the mini-batch size is set to be $\min\{1000, 0.2n\}$. 

	The three regularization parameters $(\lambda_1, \lambda_2, \lambda_3)$ are fine-tuned, as they are usually sensitive to data and model performance.  Below we provide the setting that is tested through the numerical experiments from the next section: 
	\begin{itemize}
		\item The sparsity hyperparameter $\lambda_1$: by default $10^{-3}$, and it can be tuned at log-scale within $[10^{-4},  10^{-2}]$;
		\item The sparsity hyperparameter $\lambda_2$: by default $10^{-3}$, and it can be tuned at log-scale within $[10^{-4}, 10^{-2}]$; 
		\item The smoothing hyperparameter $\lambda_3$: by default $10^{-6}$, and it can be tuned at log-scale within $[10^{-7}, 10^{-5}]$.
	\end{itemize}
	
	These hyperparameters can be selected using the hold-out validation accuracy by grid search, random search, Bayesian optimization, and many other tools in the area of automated machine learning. In this paper, we use the simple grid search and tune the first two sparsity regularization parameters, while the smoothness regularization strength is fixed to be the default value.
	
	\section{Numerical Experiments}\label{Numerical}
	In this section, we compare the proposed xNN.enhanced model to several benchmarks through simulation studies. It is shown that the intrinsically interpretable xNN.enhanced is flexible enough to approximate complex functions and achieve high prediction accuracy. We also include a real case study based on the LendingClub dataset as a showcase of the xNN.enhanced application. 
	
	\subsection{Experimental Setting}
	Several benchmark models are considered for comparison, including the xNN.naive model \citep{vaughan2018explainable}, multi-layer perceptron (MLP), extreme learning machine (ELM), support vector machine (SVM), random forest (RF), least absolute shrinkage and selection operator (LASSO) and logistic regression (LogR; only for the real data application).
	
	Each dataset is split it into training, validation and test sets, and the validation set is used to control early stopping for neural network-based models and select the best hyperparameters. In particular, we use the following settings. For LASSO, the shrinkage parameter is tuned within $\{10^{-2},10^{-1},10^{0},10^{1},10^{2}\}$. For SVM with the radial basis function (RBF) kernel, its regularization parameter and kernel parameter are selected within the grid $\{10^{-2},10^{-1},10^{0},10^{1},10^{2}\} \times \{10^{-2},10^{-1},10^{0},10^{1},10^{2}\}$. We use a two-hidden-layer MLP with 100 hyperbolic tangent nodes in the first hidden layer and 60 hyperbolic tangent nodes in the second hidden layer, respectively. For ELM, we use the version with generalized radial basis functions and random layers~\citep{huang2006extreme,fernandez2011melm}. The number of hidden nodes and kernel width parameter are tunned within $\{20, 50, 100, 200, 500\} \times \{10^{-4},10^{-3},10^{-2},10^{-1},10^{0}\}$. For RF with 100 trees, the maximum tree depth is selected from $ \{3, 4, 5, 6, 7, 8\}$. For a fair comparison, the hyperparameters for both the naive and enhanced xNN models are configured  in the same way according to the discussion in Section~\ref{Estimation:hyperpara}.	
	
	All the experiments are implemented using the Python environment on a server with multiple Intel Xeon 2.60G CPUs.  Both the naive and enhanced xNN models are implemented using the \textsl{TensorFlow 2.0} platform, while all the other benchmark models are implemented using the \textsl{Scikit-learn} and \textsl{Sklearn-extensions} packages. 
	
	\subsection{Simulation Study}  \label{Simulation}
	We consider six different scenarios under the regression settings. In all the examples, we generate 10-dimensional features  using the following mechanism: 

	\begin{enumerate} [a)]
		\item Generate the i.i.d. uniform random variables $d_{j}\sim\mbox{\rm Unif}(-1, 1)$, for $j=1,2,\ldots,10$;
		\item Generate a random variable $s$ from $\mbox{\rm Unif}(-1, 1)$;
		\item For each $j$, generate the $j$th feature by $x_{j} = \frac{d_{j}+ts}{1+t}$, where $t$ is chosen to be 1 such that the pairwise correlation 
		$$
		\rho = \mbox{\rm Corr}(x_j, x_{j'}) =  \frac{t^{2}}{1+t^{2}} = 0.5, \mbox{ for } j\neq j'. 
		$$
		
	\end{enumerate}
	Then, the response $y$ is generated according to different model assumptions as specified below. 
	
	\medskip			
	\ni \textbf{S1: Additive model (orthogonal projection)} \\ This is a case that follows the enhanced xNN model assumption. It consists of four additive components with mutually orthogonal projection indexes, i.e.,	
	\begin{equation}\label{S1}
	\begin{split}
	y = \sum_{j=1}^4 h_{j}(\bm{w}_j^{T}\bm{x})  +\varepsilon,
	\end{split}
	\end{equation}
	where the weights and ridge functions are given by 
	$$
	\begin{gathered}
	\bm{W}^{T} = 
	\begin{bmatrix}
	1 & 0 & 0   & 0   & 0   & 0   & 0   & 0 & 0 & 0 \\
	0 & 1 & 0   & 0   & 0   & 0   & 0   & 0 & 0 & 0 \\
	0 & 0 & 0.5 & 0.5 & 0   & 0   & 0   & 0 & 0 & 0 \\
	0 & 0 & 0   & 0   & 0.2 & 0.3 & 0.5 & 0 & 0 & 0
	\end{bmatrix} , \\
	h_1(z) = 2z, \; h_{2}(z) = 0.2e^{-4z},\; \\
	h_{3}(z) = 3z^2, \; h_{4}(z) = 2.5\sin(\pi z). 
	\end{gathered} 
	$$
	Note that in the $\W$ matrix, the last three features are treated as inactive variables.
	
	\medskip
	\ni \textbf{S2: Additive model (near-orthogonal projection)} \\
	In this scenario, the model also takes the additive form, while the projection indexes are not mutually orthogonal, 
	\begin{equation}\label{S2}
	y = 3 + h_{1}(\bm{w}_{1}^{T}\bm{x}) + h_{2}(\bm{w}_{2}^{T}\bm{x})+ h_{3}(\bm{w}_{3}^{T}\bm{x}) + \varepsilon,
	\end{equation}
	where the weights and ridge functions are given by
	$$
	\begin{gathered}
	\bm{W}^{T} = 
	\begin{bmatrix}
	0.1 & 0.9 & 0   & 0   & 0 & 0 & 0 & 0 & 0 & 0 \\
	0   & 0.1 & 0.9 & 0   & 0 & 0 & 0 & 0 & 0 & 0 \\
	0   & 0   & 0.1 & 0.9 & 0 & 0 & 0 & 0 & 0 & 0
	\end{bmatrix} , \\
	h_{1}(z) = 0.5z, \; h_{2}(z) =\frac{4\sin(\pi z)}{2-\sin(\pi z)} , \; h_{3}(z) = -4\exp(-z^2).
	\end{gathered} 
	$$
	
	\medskip
	\ni \textbf{S3 -- S6: Non-additive models}\\
	Consider the following four models that all violate the AIM assumption: 
	\begin{equation}\label{S3}
	y = e^{2\tanh (x_{1}x_{2} + 2x_{3}x_{4})} + \varepsilon,
	\end{equation}
	\begin{equation}\label{S4}
	y = 3\pi^{x_{1}x_{2}} \sqrt{2(x_{3}+1)} + \varepsilon,
	\end{equation}
	\begin{equation}\label{S5}
	y = x_{1} - x_{2} + \frac{2(x_{3}+x_{4}+x_{5}+x_{6})}{0.5+(1.5+x_{3}+x_{5}-x_{4}-x_{6})^2} +\varepsilon,
	\end{equation}	
	\begin{equation}\label{S6}
	y =  \sin\left[\frac{1}{2}\pi(-x_{1}+2x_{3}+x_{4})\right]e^{\frac{1}{2}(x_{2}+x_{3}-x_{4})} + \varepsilon.
	\end{equation}
	
	In all the six scenarios, we set the white noise $\varepsilon \sim N(0,1)$. Four different sample sizes are considered, including $n = \{1000, 2000, 5000, 10000\}$. Each data is further split into two parts, with 80{\%} for training and 20{\%} for validation. Moreover, a test set with $10000$ samples is generated following the same data generation mechanism.

	\begin{table}[!t]
		\centering
		\renewcommand\tabcolsep{3pt}
		\renewcommand\arraystretch{1.2}
		\caption{Testing MSE comparison under Scenarios 1--6.} \label{Simu}
		\begin{tabular}{cc|ccccccc}
			\hline
			\raisebox{-2ex}{Name} & \raisebox{-2ex}{Size} &    \raisebox{-1ex}{xNN}     &     \raisebox{-1ex}{xNN}      & \raisebox{-2ex}{SVR} & \raisebox{-2ex}{RF} & \raisebox{-2ex}{MLP} & \raisebox{-2ex}{ELM} & \raisebox{-2ex}{LASSO} \\
			&    &  \raisebox{1ex}{(enhanced)}   &   \raisebox{1ex}{(naive)}      &    &     &    &    & \\ \hline
			
			S1            &           1000           & \textbf{1.086} &     1.088      &          1.274          &         1.415          &          1.232          &          1.476          &           2.584           \\
			S1            &           2000           & \textbf{1.039} &     1.043      &          1.168          &         1.308          &          1.121          &          1.260          &           2.548           \\
			S1            &           5000           & \textbf{1.007} &     1.009      &          1.090          &         1.251          &          1.043          &          1.149          &           2.561           \\
			S1            &          10000           & \textbf{1.004} &     1.005      &          1.069          &         1.239          &          1.023          &          1.091          &           2.600           \\ \hline\hline
			S2            &           1000           &     1.068      & \textbf{1.065} &          1.460          &         1.189          &          1.134          &          1.622          &           2.141           \\
			S2            &           2000           & \textbf{1.030} &     1.031      &          1.305          &         1.119          &          1.066          &          1.384          &           2.131           \\
			S2            &           5000           & \textbf{1.002} & \textbf{1.002} &          1.169          &         1.067          &          1.019          &          1.254          &           2.116           \\
			S2            &          10000           &     1.002      & \textbf{1.000} &          1.109          &         1.054          &          1.011          &          1.167          &           2.129           \\ \hline\hline
			S3            &           1000           &     1.100      & \textbf{1.099} &          1.109          &         1.127          &          1.115          &          1.177          &           1.370           \\
			S3            &           2000           & \textbf{1.038} &     1.065      &          1.060          &         1.092          &          1.058          &          1.108          &           1.362           \\
			S3            &           5000           & \textbf{1.005} &     1.019      &          1.022          &         1.057          &          1.018          &          1.057          &           1.349           \\
			S3            &          10000           & \textbf{1.003} &     1.009      &          1.016          &         1.057          &          1.012          &          1.037          &           1.361           \\ \hline\hline
			
			S4           &           1000           & \textbf{1.095} &     1.135      &          1.196          &         1.280          &          1.220          &          1.356          &           2.121           \\
			S4            &           2000           & \textbf{1.052} &     1.068      &          1.127          &         1.182          &          1.140          &          1.208          &           2.098           \\
			S4            &           5000           &     1.032      & \textbf{1.024} &          1.065          &         1.110          &          1.058          &          1.103          &           2.087           \\
			S4            &          10000           & \textbf{1.030} &     1.035      &          1.045          &         1.089          &     \textbf{1.030}      &          1.060          &           2.118           \\ \hline\hline
			S5            &           1000           &     1.069      & \textbf{1.062} &          1.074          &         1.154          &          1.069          &          1.145          &           1.243           \\
			S5            &           2000           &     1.045      & \textbf{1.030} &          1.041          &         1.101          &          1.041          &          1.081          &           1.237           \\
			S5            &           5000           &     1.013      & \textbf{1.003} &          1.015          &         1.060          &          1.014          &          1.029          &           1.231           \\
			S5            &          10000           &     1.010      & \textbf{1.001} &          1.012          &         1.052          &          1.011          &          1.020          &           1.240           \\ \hline\hline
			S6            &           1000           &     1.105      & \textbf{1.083} &          1.187          &         1.168          &     \textbf{1.083}      &          1.292          &           1.300           \\
			S6            &           2000           &     1.058      & \textbf{1.040} &          1.141          &         1.107          &          1.047          &          1.203          &           1.300           \\
			S6            &           5000           &     1.027      & \textbf{1.003} &          1.074          &         1.064          &          1.018          &          1.105          &           1.289           \\
			S6            &          10000           &     1.027      & \textbf{1.003} &          1.054          &         1.050          &          1.012          &          1.078          &           1.287           \\ \hline\hline
		\end{tabular}
	\end{table}	
	
	Table~\ref{Simu} summarizes the results of the compared models under Scenarios 1 to 6, respectively. The experiments are repeated for 10 times with mean square error (MSE) on the test set being reported. In particular, the best results are highlighted in bold. Even though Scenarios 2 - 6 are generated by functions that do not follow the assumption of (\ref{SOSxNN}), the xNN.enhance model still achieves the best or nearly the best performance in most cases. Therefore, we can claim that the proposed xNN model is competitive regarding prediction accuracy.
	
\begin{figure*}[!t]
		\centering
		\subfloat[Ground Truth]{
			\label{Simu1_GT} 
			\includegraphics[width=0.32\textwidth, height = 0.33\textheight]{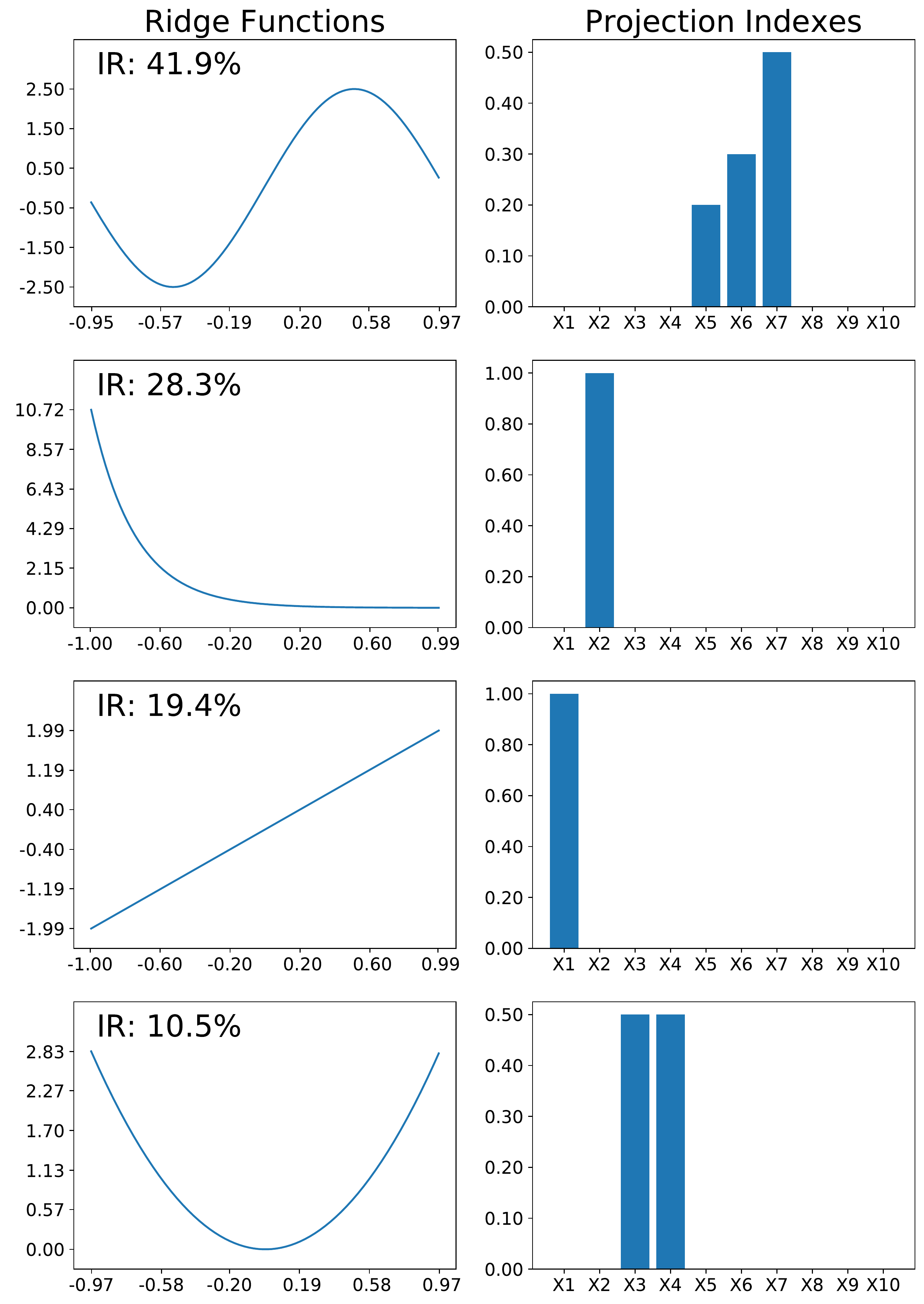}}
		\subfloat[t][xNN.naive]{
			\label{Simu1_xNN} 
			\includegraphics[width=0.32\textwidth, height = 0.41\textheight]{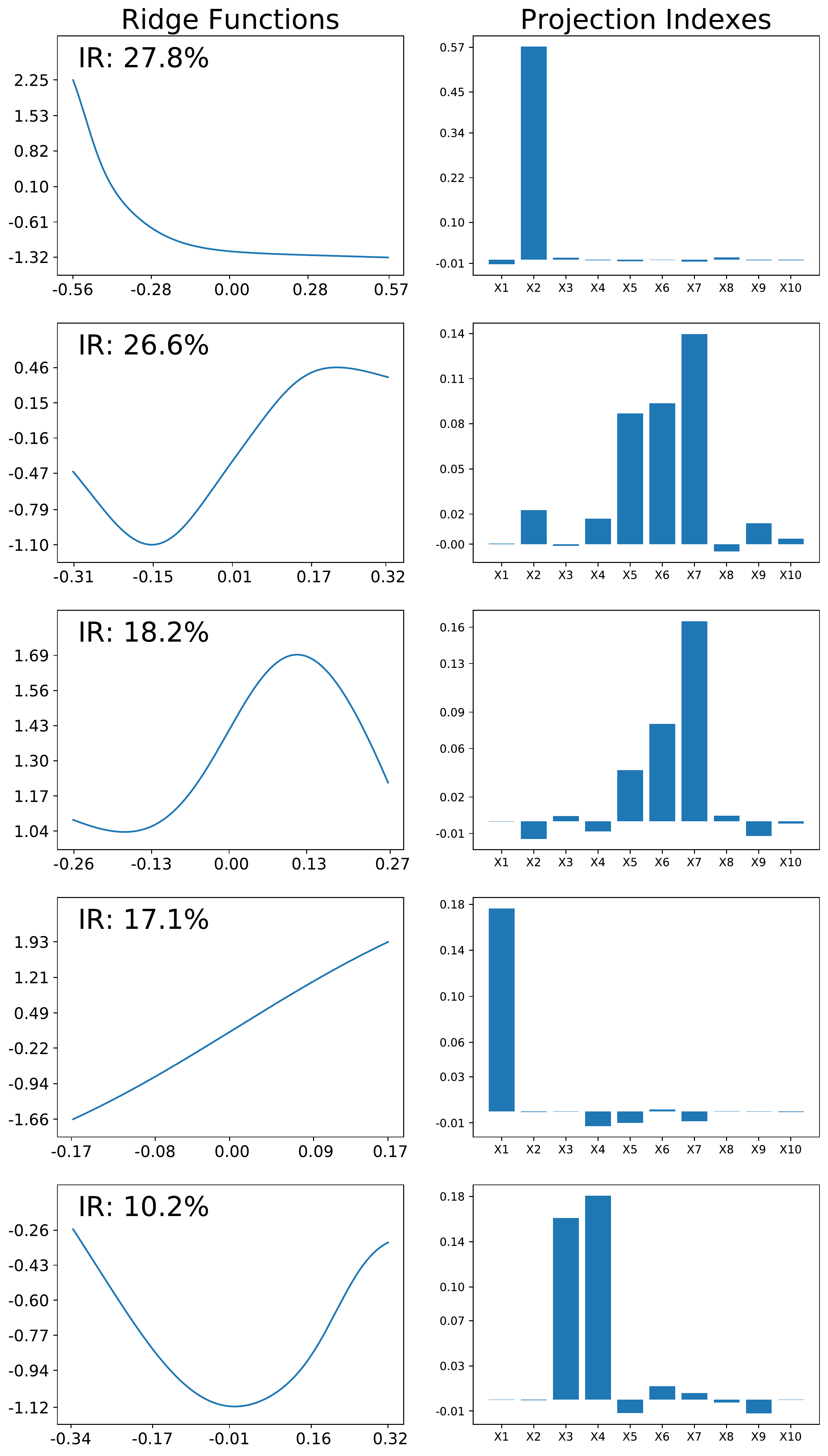}}
		\subfloat[xNN.enhanced]{
			\label{Simu1_SOSxNN} 
			\includegraphics[width=0.32\textwidth, height = 0.33\textheight]{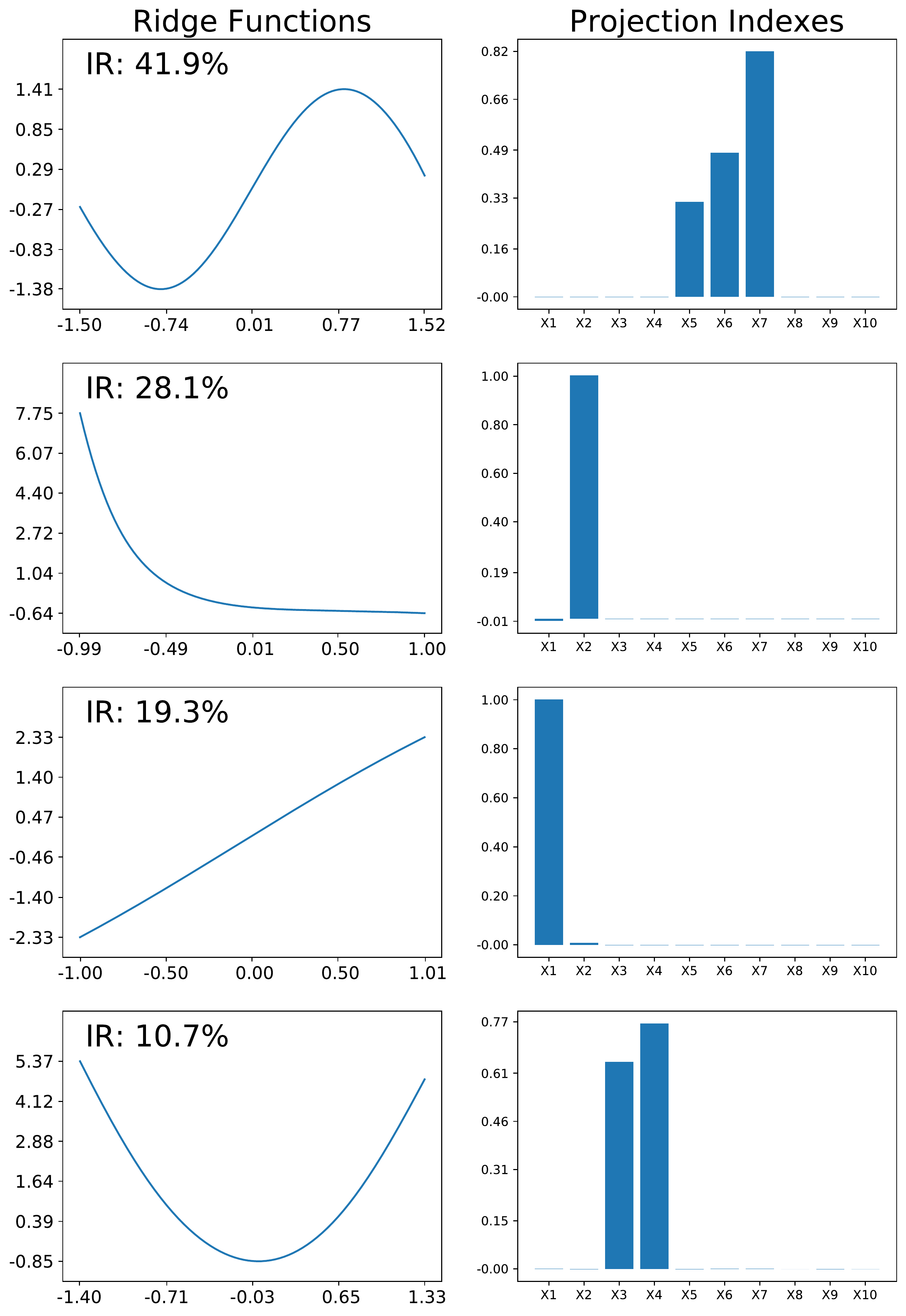}}
		\caption{Visualized model fits (vs. the ground truth) for Scenario 1. }\label{Simu1_Visu}
		
		\bigbreak
		\subfloat[Ground Truth]{
			\label{Simu2_GT} 
			\includegraphics[width=0.32\textwidth, height = 0.27\textheight]{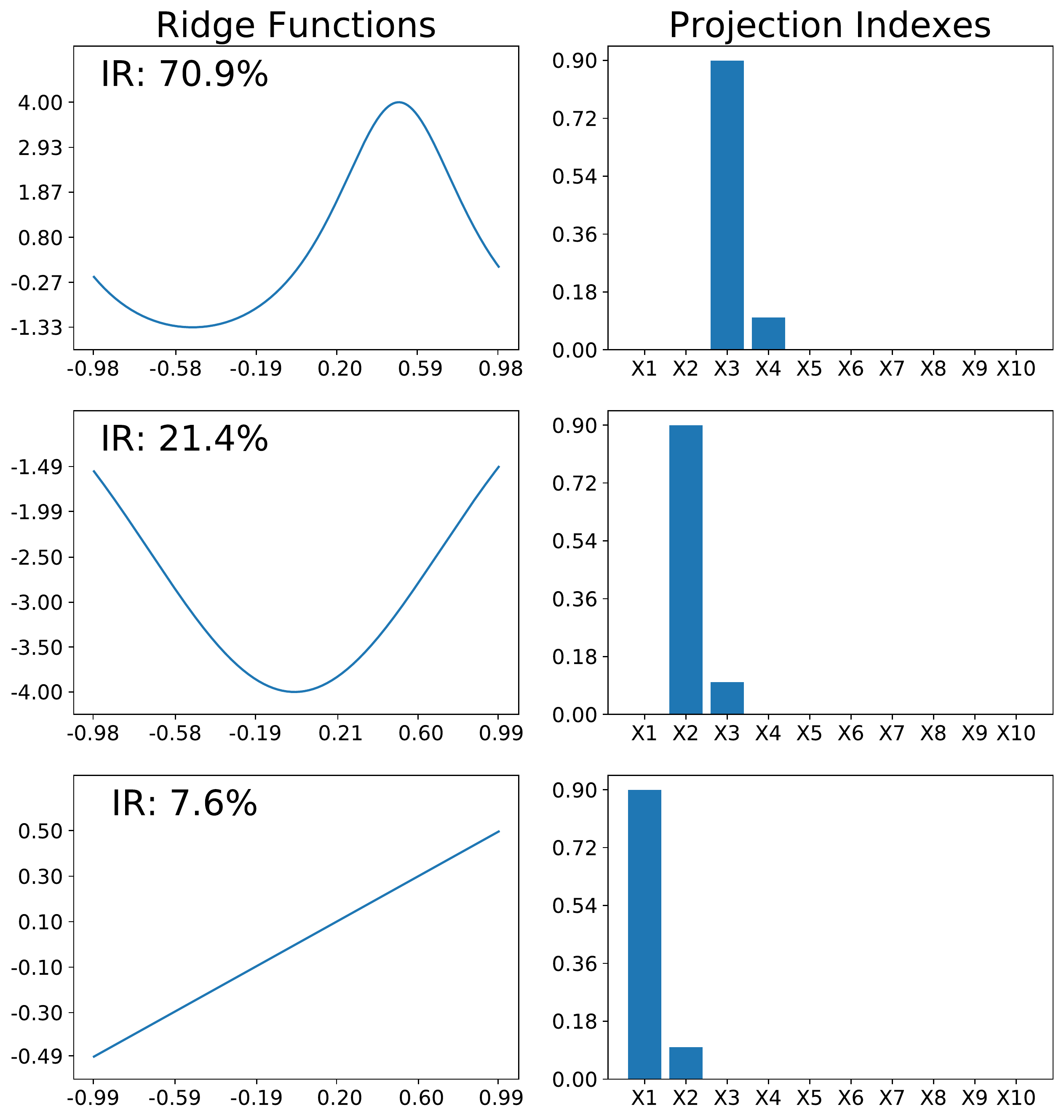}}
		\subfloat[xNN.naive]{
			\label{Simu2_xNN} 
			\includegraphics[width=0.32\textwidth, height = 0.27\textheight]{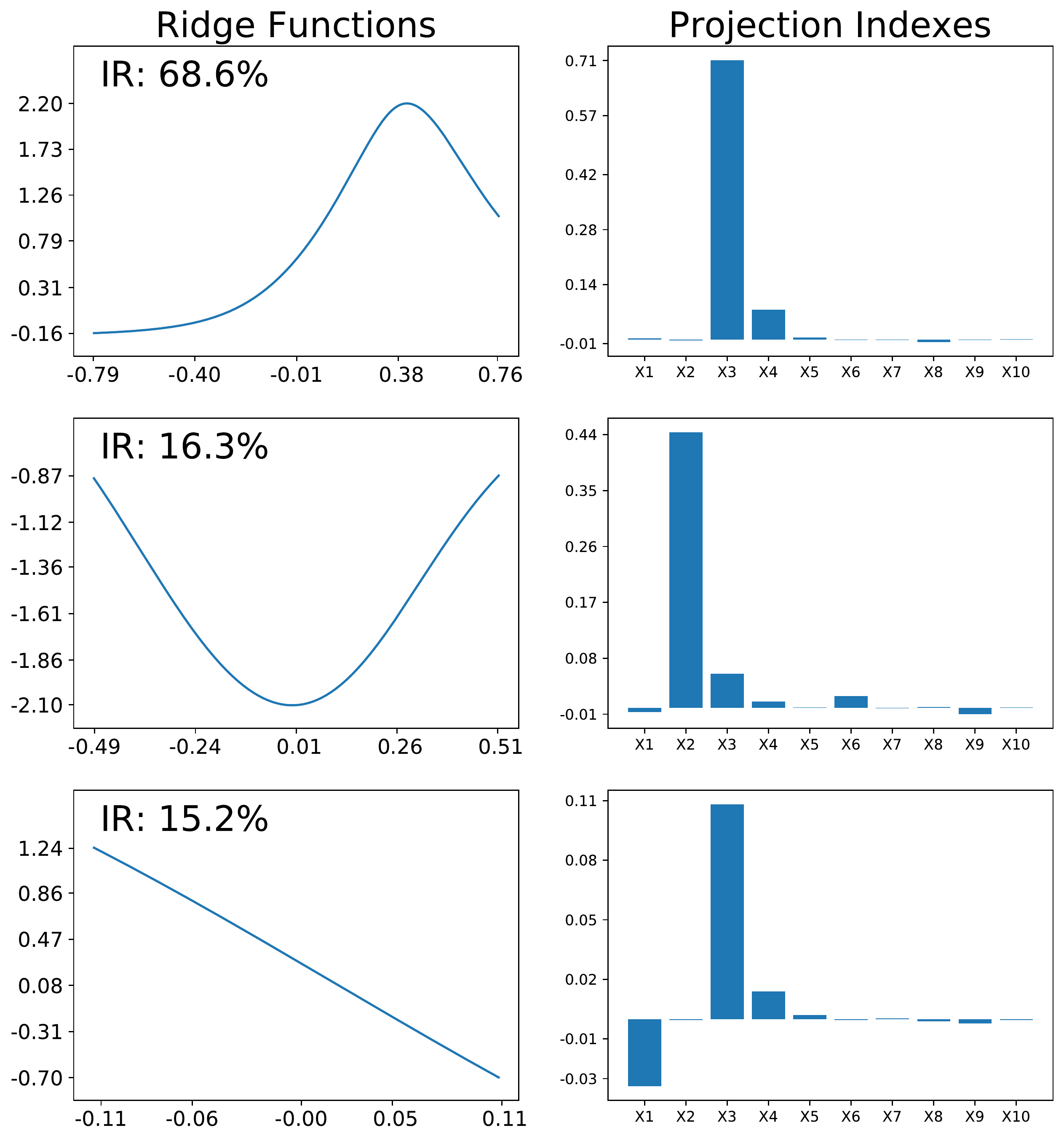}}
		\subfloat[xNN.enhanced]{
			\label{Simu2_SOSxNN} 
			\includegraphics[width=0.32\textwidth, height = 0.27\textheight]{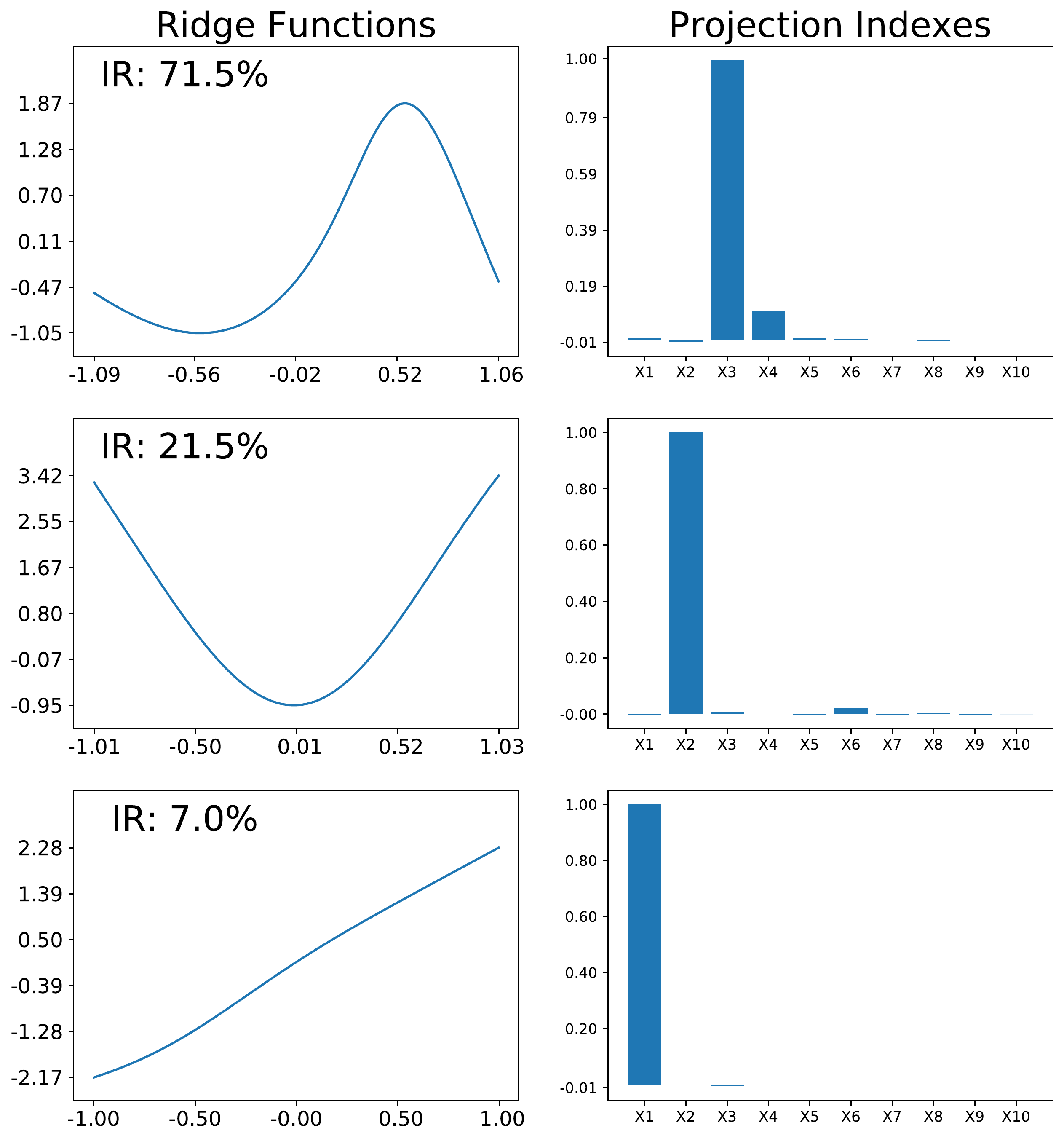}} 
		\caption{Visualized model fits (vs. the ground truth) for Scenario 2. }\label{Simu2_Visu}
	\end{figure*}

\begin{figure*}[!t]
		\centering	
		\subfloat[t][xNN.naive]{
			\label{Simu3_xNN} 
			\includegraphics[width=0.4\textwidth, height = 0.212\textheight]{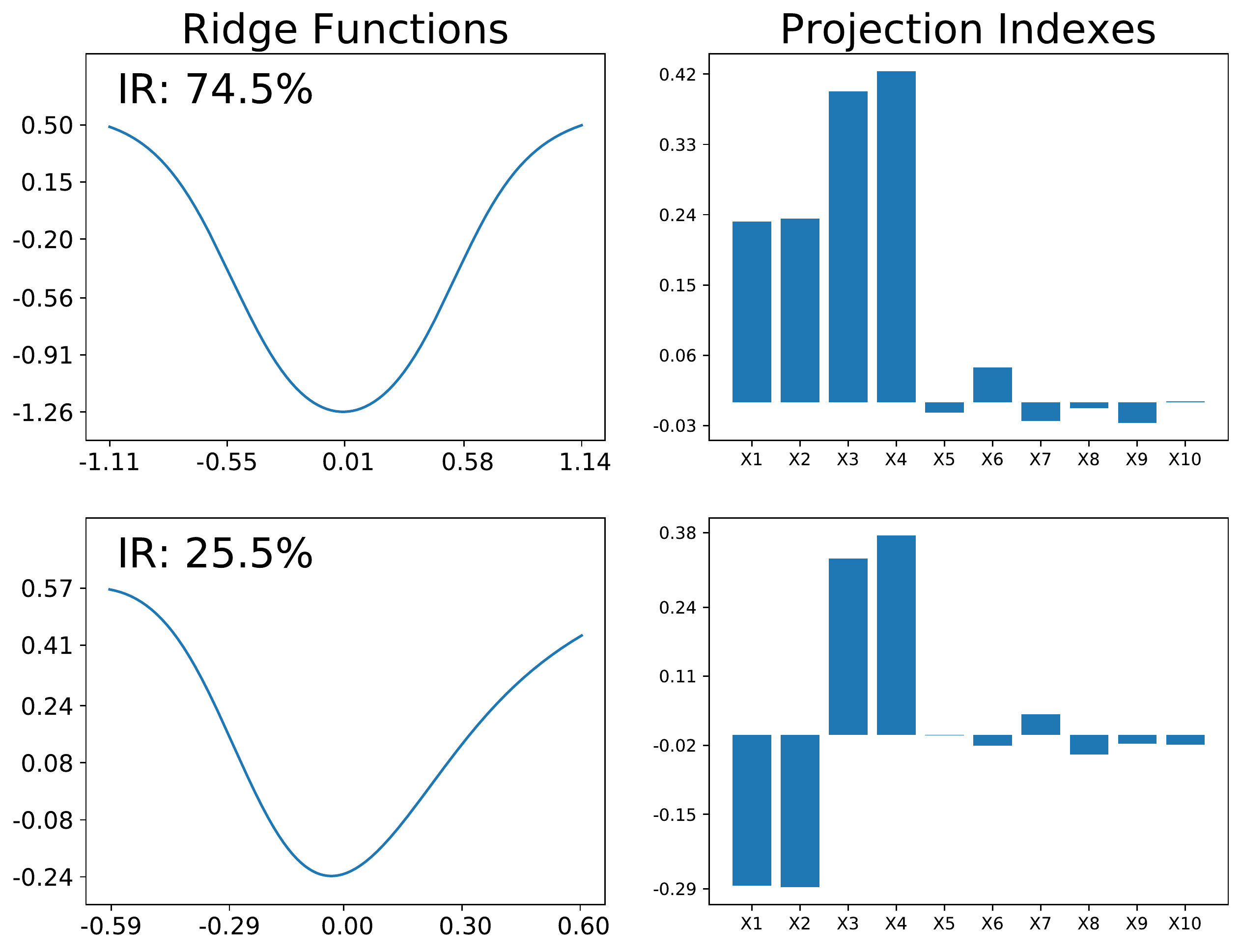}}
		\smallskip 
		\subfloat[t][xNN.enhanced]{
			\label{Simu3_SOSxNN} 
			\includegraphics[width=0.4\textwidth]{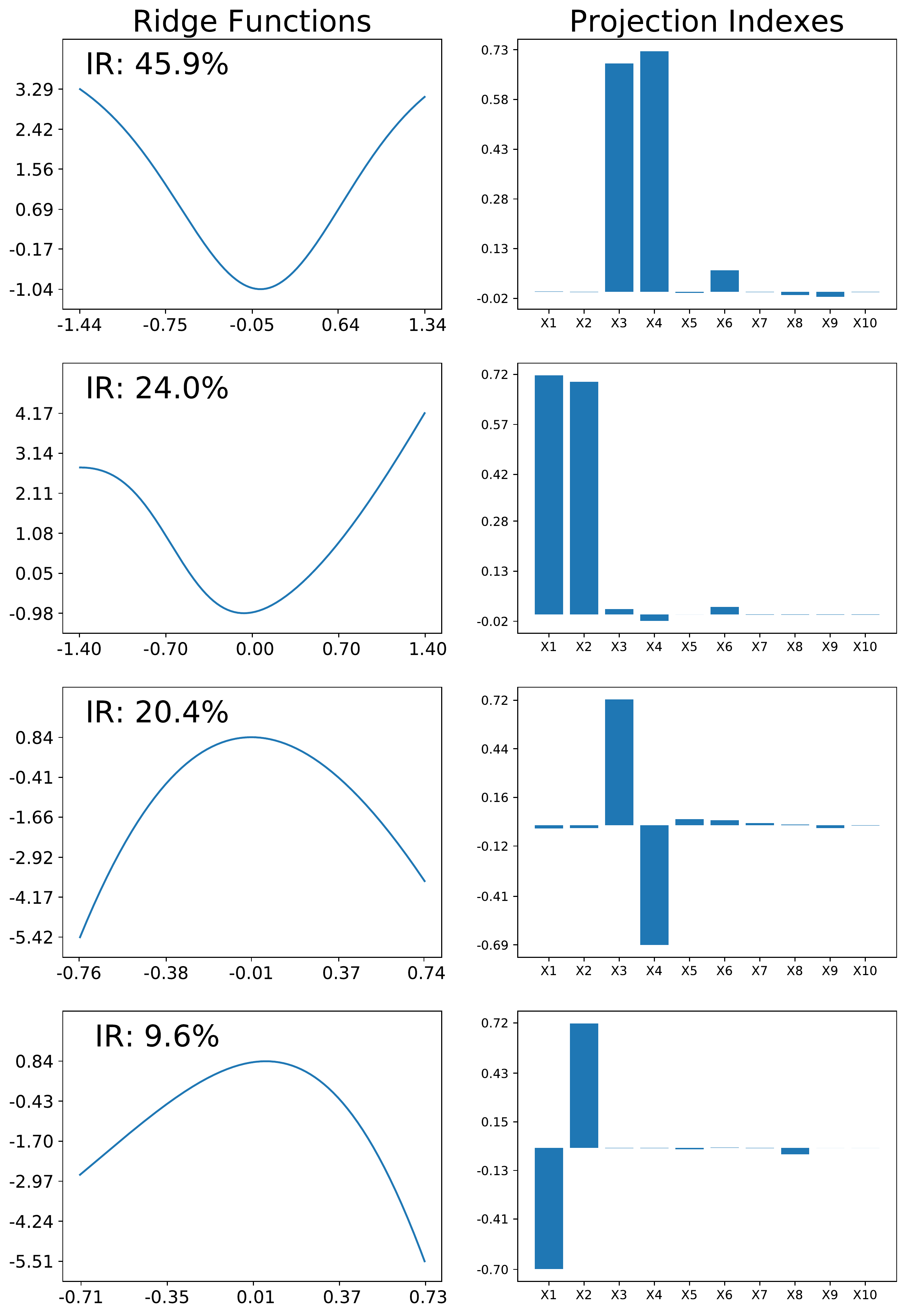}}
		\caption{Visualized model fits for Scenario 3.}\label{Simu3_Visu}
	\end{figure*}

\begin{figure*}[!t]
		\centering	
		\subfloat[t][xNN.naive]{
			\label{Simu4_xNN} 
			\includegraphics[width=0.4\textwidth]{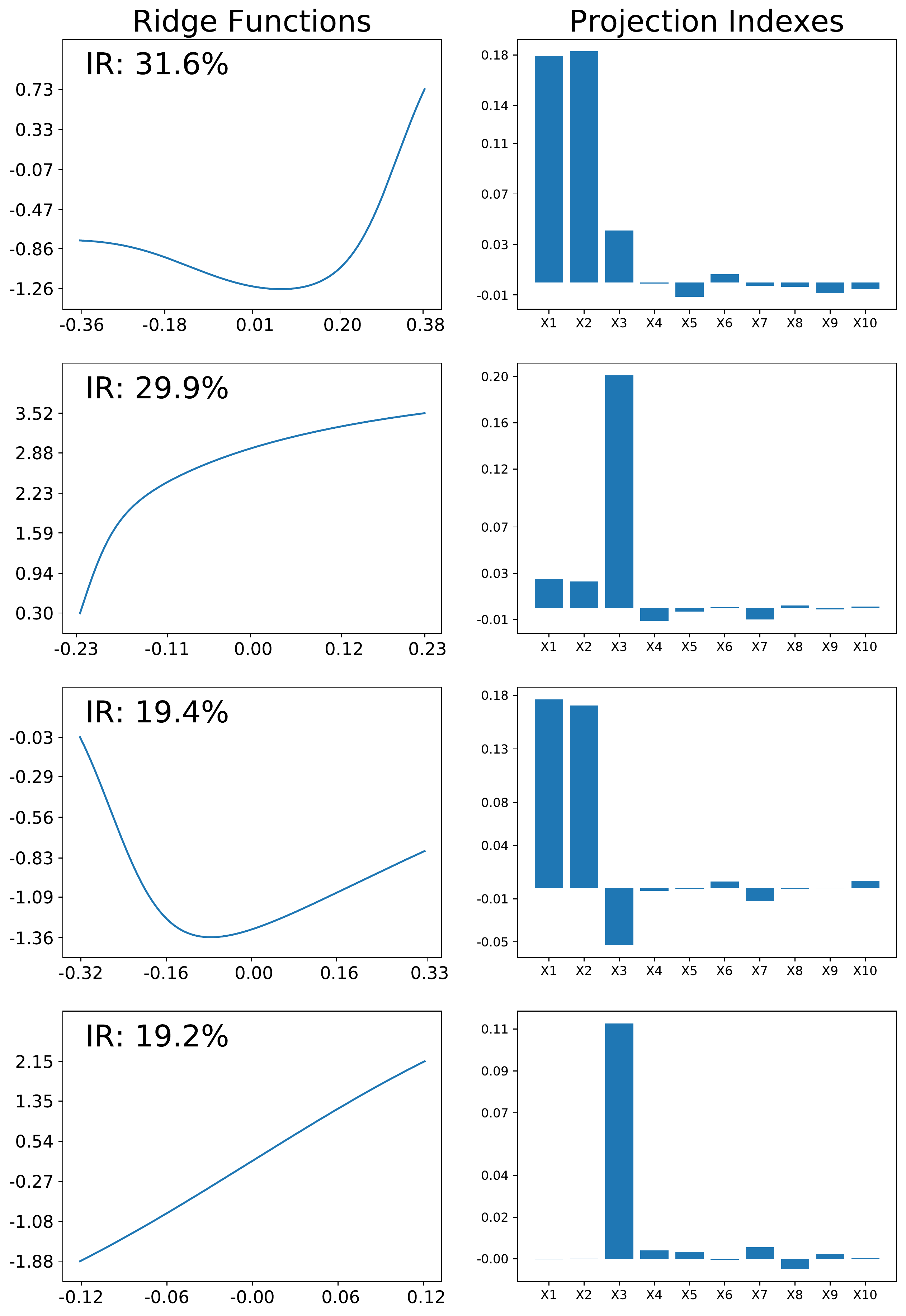}}
		\smallskip 
		\subfloat[t][xNN.enhanced]{
			\label{Simu4_SOSxNN} 
			\includegraphics[width=0.4\textwidth, height = 0.315\textheight]{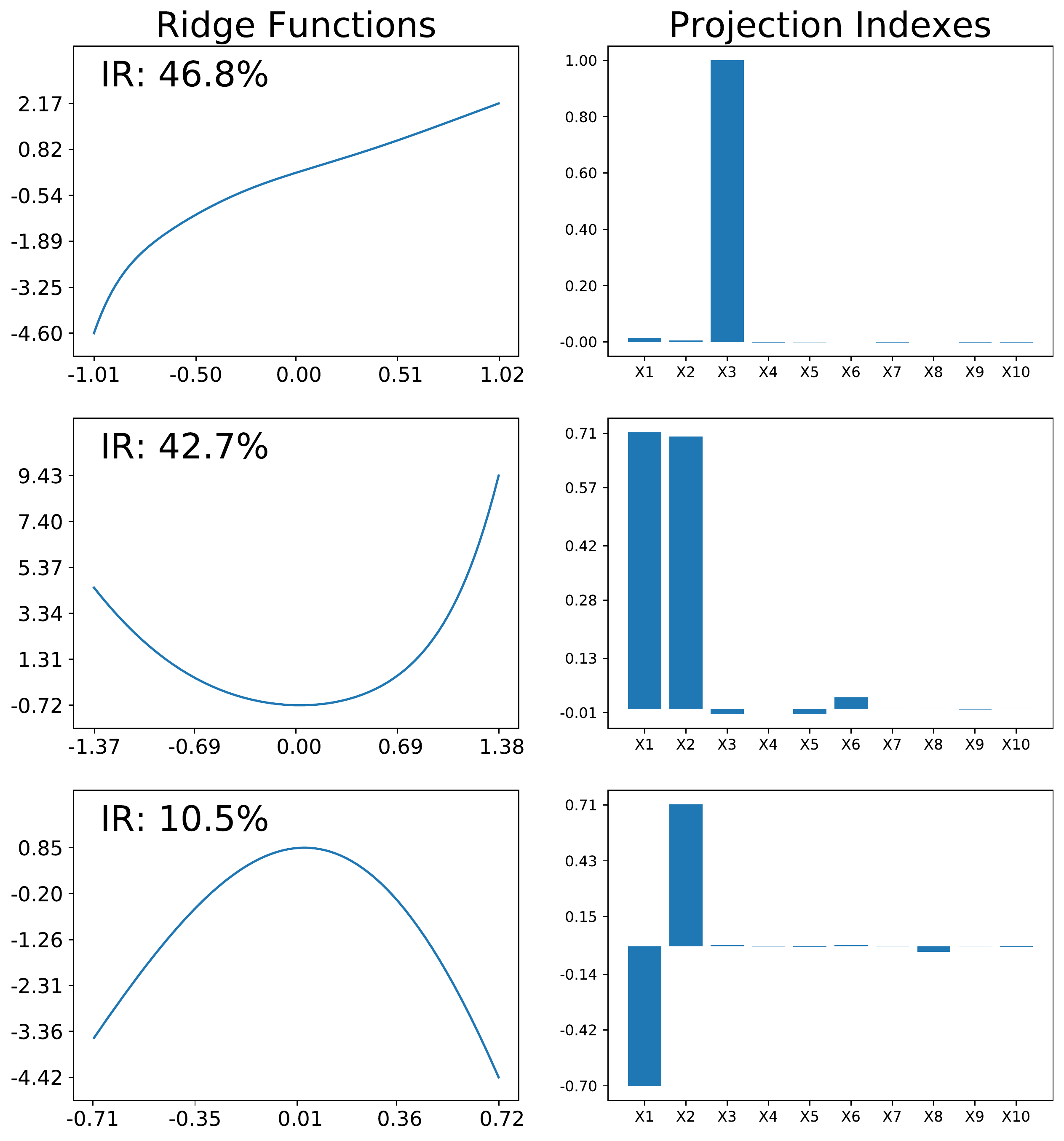}}
		\caption{Visualized model fits for Scenario 4.}\label{Simu4_Visu}
	\end{figure*}

\begin{figure*}[!t]
		\centering	
		\subfloat[t][xNN.naive]{
			\label{Simu5_xNN} 
			\includegraphics[width=0.4\textwidth, height = 0.315\textheight]{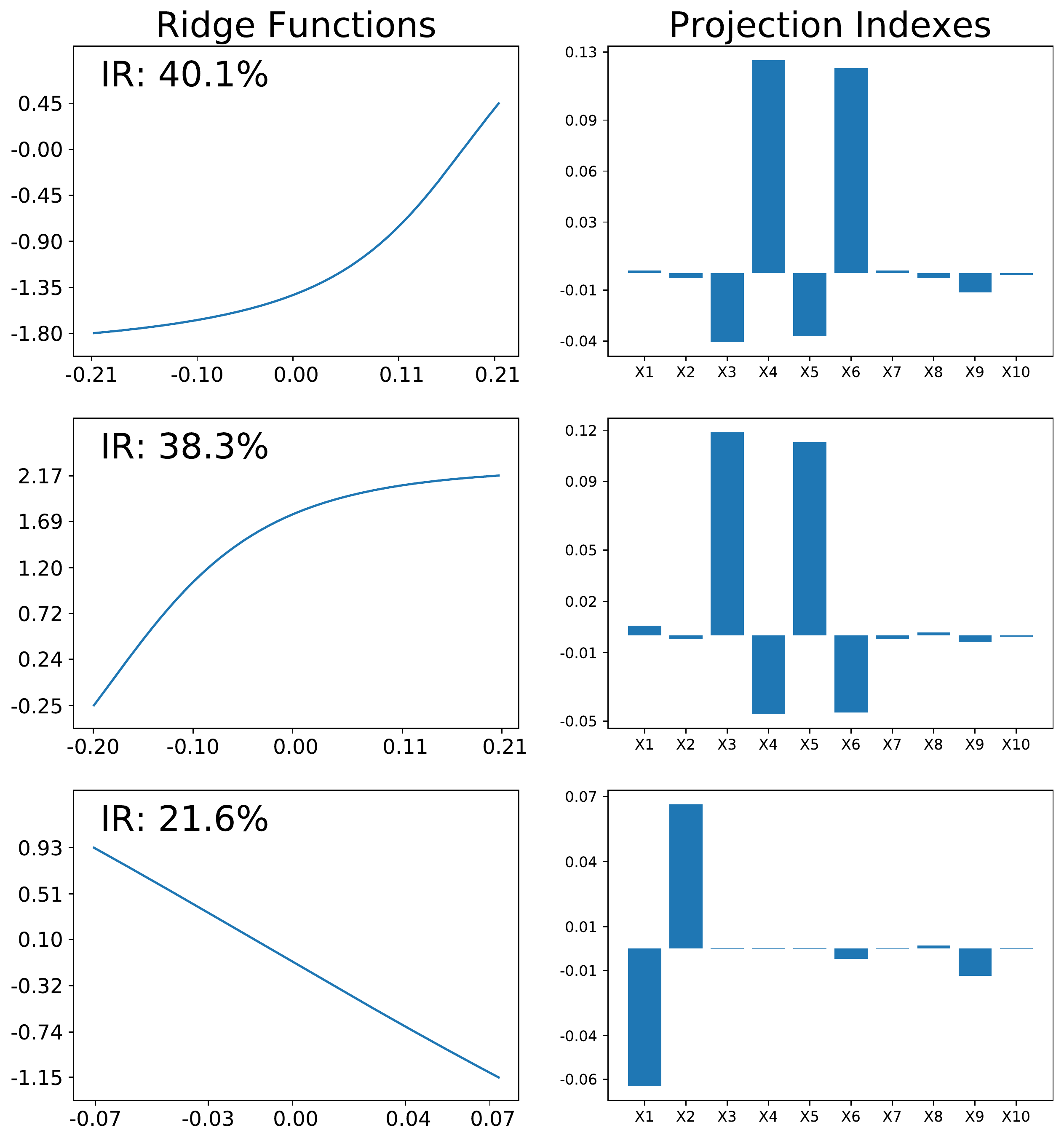}}
		\smallskip 
		\subfloat[t][xNN.enhanced]{
			\label{Simu5_SOSxNN} 
			\includegraphics[width=0.4\textwidth]{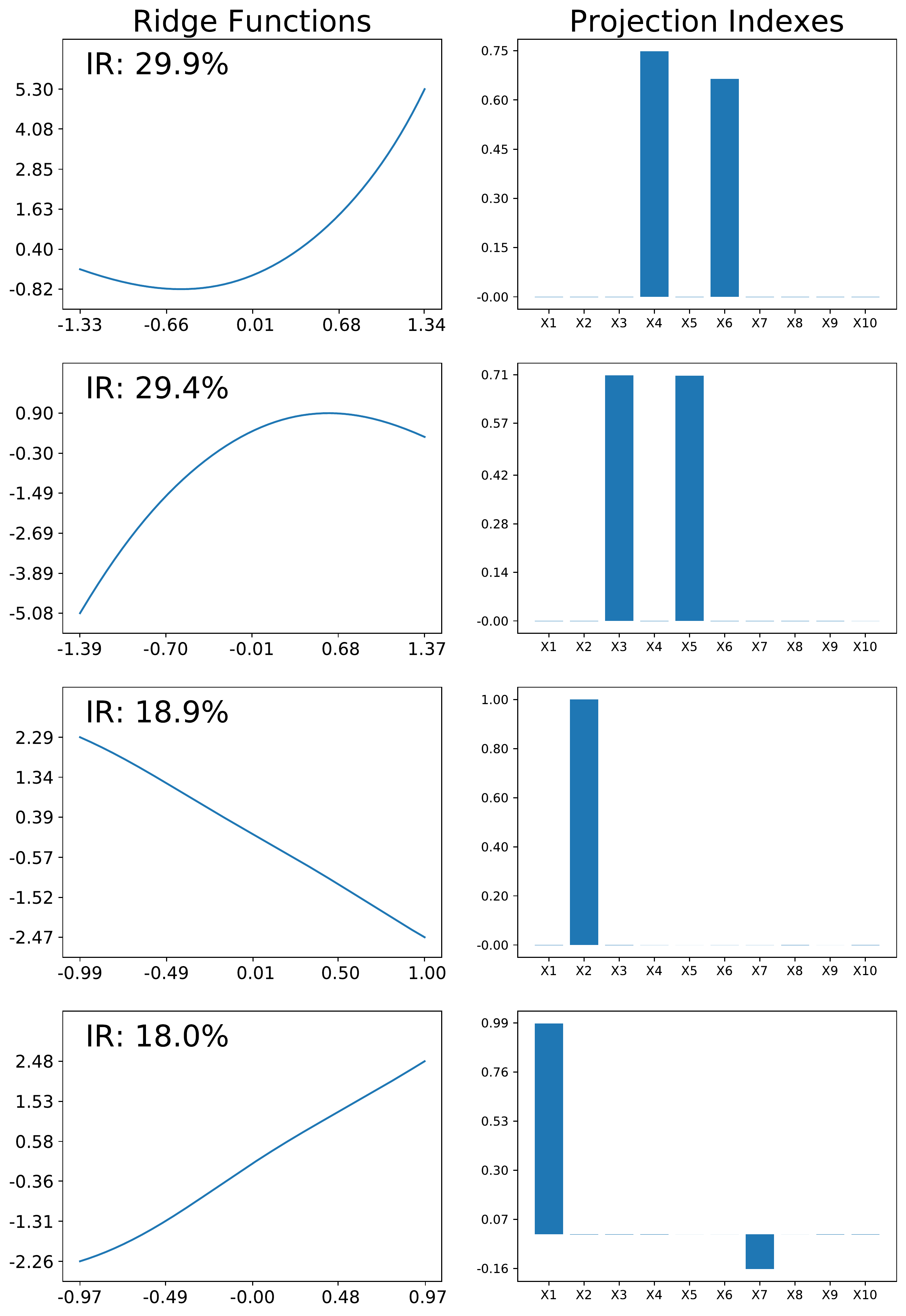}}
		\caption{Visualized model fits for Scenario 5.}\label{Simu5_Visu}
	\end{figure*}

\begin{figure*}[!t]
		\centering
		\subfloat[t][xNN.naive]{
			\label{Simu6_xNN} 
			\includegraphics[width=0.4\textwidth]{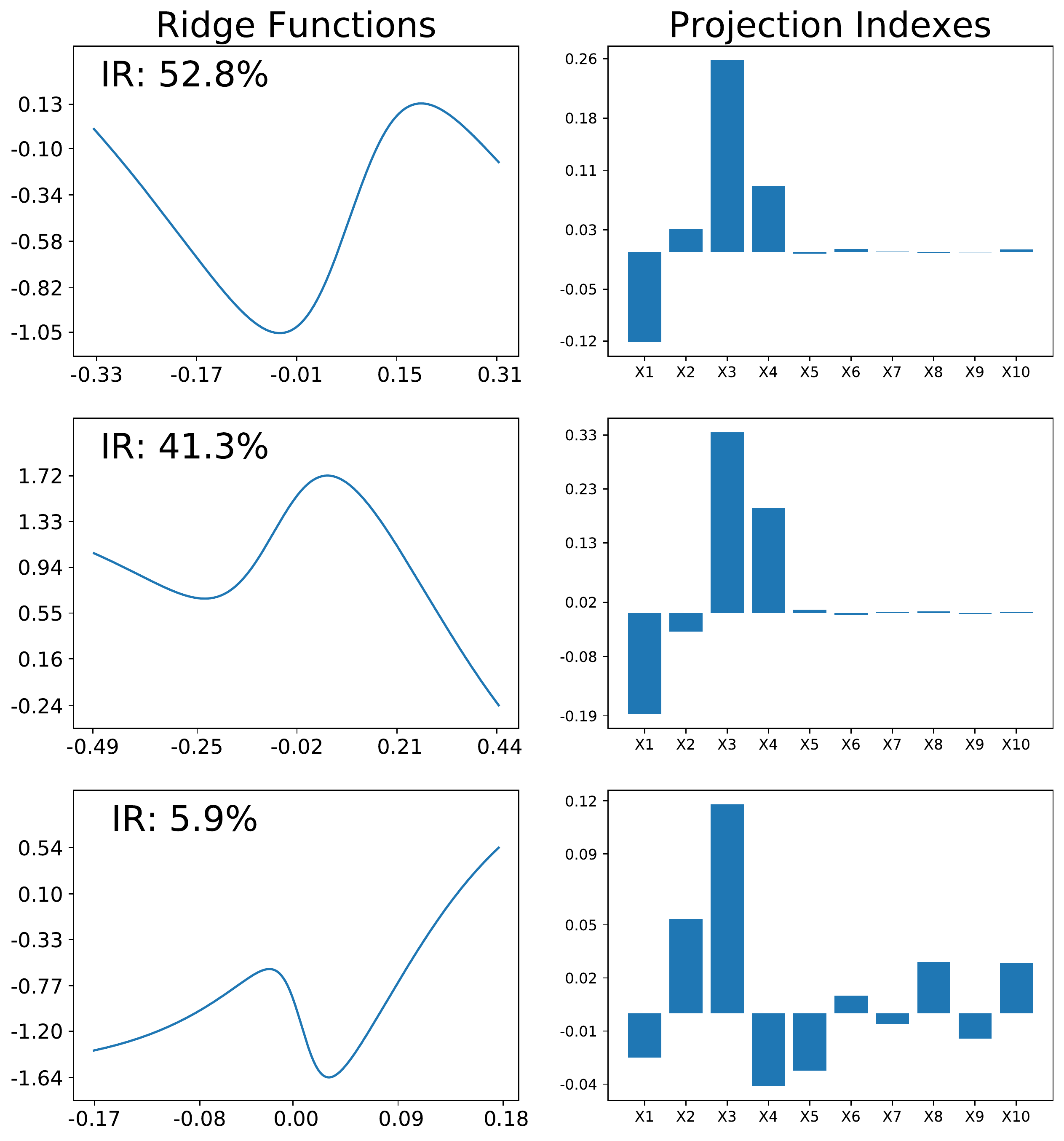}}
		\smallskip 
		\subfloat[t][xNN.enhanced]{
			\label{Simu6_SOSxNN} 
			\includegraphics[width=0.4\textwidth]{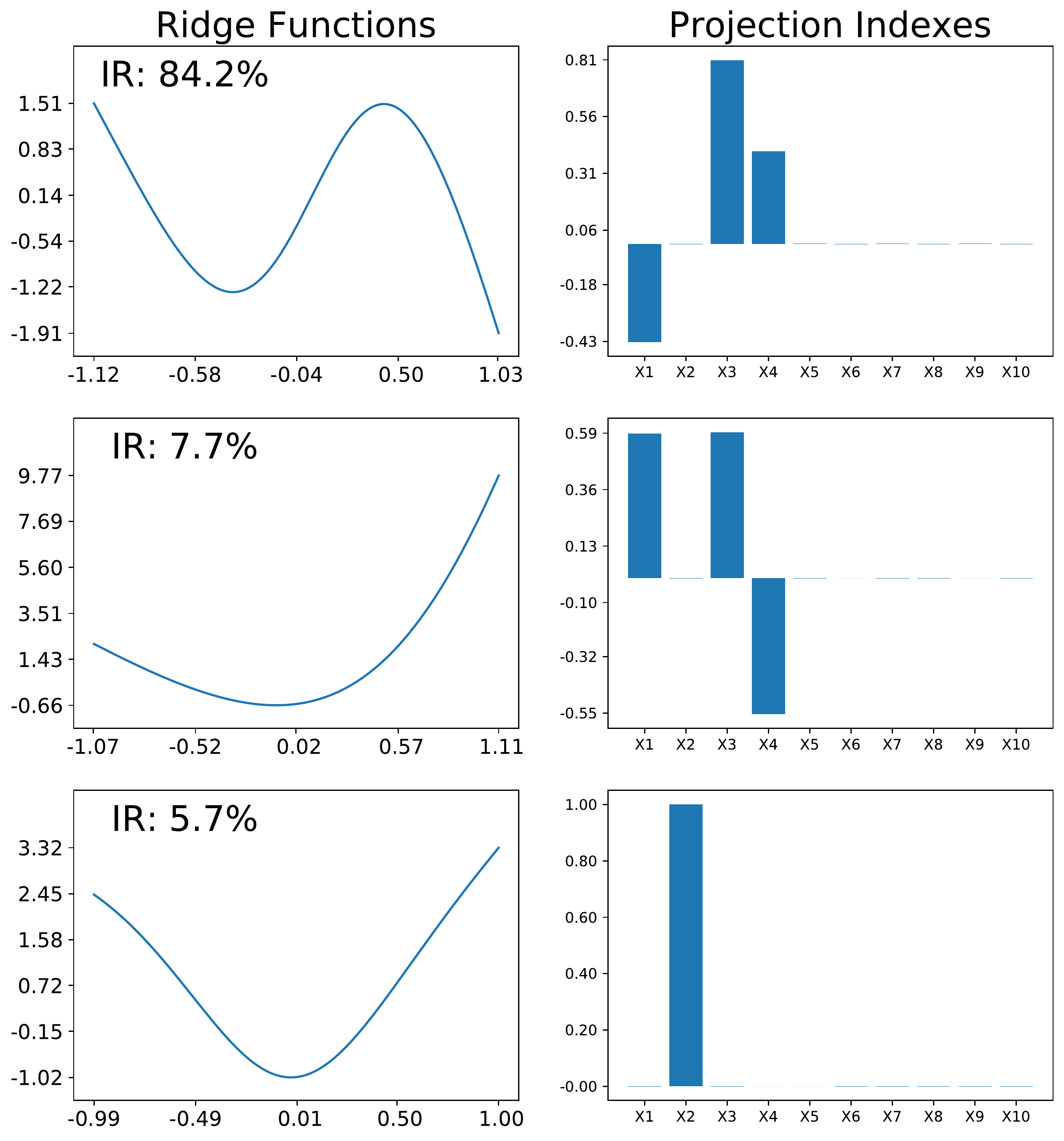}}
		\caption{Visualized model fits for Scenario 6.}\label{Simu6_Visu}
	\end{figure*}

	The proposed xNN model with enhanced explainability can be checked from Figs.~\ref{Simu1_Visu}--\ref{Simu6_Visu}. We draw the ridge functions and projection indexes for both the naive and enhanced xNN model estimates, with one repetition of 10000 samples (Note the estimated results may be slightly different among the repetitions). For Scenarios 1 and 2, the left, middle, and right panels of Figs.~\ref{Simu1_Visu}--\ref{Simu2_Visu} represent the ground truth, the xNN.naive estimation, and the xNN.enhanced estimation, respectively. For Scenarios 3-6, the corresponding ground truth plot is not available, since the underlying functions do not follow the form of (\ref{AIM}). For each sub-figure, the left panel shows the ridge function, and the right panel presents the corresponding projection index. Moreover, the ridge functions and corresponding projection indexes are sorted in the descending order of the importance ratio (IR).
	
	In Scenario 1, the sine curve $h_{4}(z)$ is the dominant component, followed by the exponential curve $h_{2}(z)$, the linear curve $h_{1}(z)$ and the quadratic term $h_{3}(z)$. It can be seen in Fig.~\ref{Simu1_Visu} that the four ridge functions and corresponding projections are all successfully recovered by the enhanced xNN, which shows superior performance over the xNN.naive model. More specifically, the xNN.naive model suffers from correlated projection indexes, where the original sine curve is represented by two similar components. With the orthogonality constraint, such a problem can be efficiently avoided, which leads to improved model interpretability. 
	
	Scenario 2 is designed with mutually near-orthogonal projection indexes. Interestingly, the enhanced xNN model can still find a good approximation to the ground truth. As shown in Fig.~\ref{Simu2_Visu}, the main effect $h_{2}(z)$ is captured with almost correct ridge function and projection indexes. The less important $h_{1}(z)$ and $h_{3}(z)$ are also well approximated. Referring to Table~\ref{Simu}, we can see such an approximation generalizes well on the test set as compared to the benchmark models. In contrast, the estimation results by xNN.naive still suffer from identifiability issues. That is, the first component is  highly confounded with the third component, where the original linear term of $x_{1}$ is mixed into other nonlinear terms.

	Although Scenarios 3 - 6 are all complex functions which do not have the additive form, we can visualize their approximations for interpretation, as shown in Figs.~\ref{Simu3_Visu} -- \ref{Simu6_Visu}. Consistent with our previous findings, the naive xNN model tends to have highly correlated and non-sparse projections, which makes the estimated xNN.naive model hard to interpret. The xNN.enhanced model is free from such problems, and thus it is generally more explainable. For example, in Fig.~\ref{Simu3_SOSxNN}, two groups of quadratic terms are captured, which are close to the forms of $ x_{1}x_{2} = \frac{1}{4}\left[(x_{1}+x_{2})^2 - (x_{1} - x_{2})^2\right] $ and $x_{3}x_{4} = \frac{1}{4}\left[(x_{3}+x_{4})^2 - (x_{3} - x_{4})^2\right]$. In Fig.~\ref{Simu4_SOSxNN}, similar results are observed for ($x_{1}, x_{2}$), and the relationship between $x_{3}$ and the response looks like a square root function. In Fig.~\ref{Simu5_SOSxNN}, xNN.enhanced reveals that ($x_{3}, x_{5}$) and ($x_{4}, x_{6}$) are two different groups of variables with nonlinear ridge functions; the variables $x_{1}$ and $x_{2}$ are instead linearly related to the response. In Fig.~\ref{Simu6_SOSxNN}, it can be seen that the importance ratio of the sine-like curve is greater than 80\%, subject to the projection of $(x_{1}, x_{3}, x_{4})$. The remaining effects are further explained by the rest components whose projection indexes are mutually orthogonal.

	\subsection{Real Data Application} \label{Lending_club}
	The peer-to-peer (P2P) lending is a method of lending money through online services by matching individual lenders and borrowers. It has been one of the hottest FinTech applications. The LendingClub dataset is obtained from (\url{https://www.lendingclub.com/ info/download-data.action}), including all the issued loans and declined loan applications that do not meet LendingClub credit underwriting policy, from Jan. 2015 to Jun. 2018. Each sample represents a loan application with six features, and the binary flag indicates the approval result (approved or declined). In particular, a data cleaning procedure is implemented to remove samples with missing values or outliers. We delete the cases whose risk score is greater than 850, as that is out of the range of FICO score. Moreover, samples with debt-income-ratio outside the range of 0 -- 200\% or with a requested amount greater than 40000 are removed. As a result, the preprocessed dataset has 1,433,570 accepted applications, and 5,607,986 declined cases. Such a dataset is then split into three parts, with 40\% for training, 10\% for validation and the rest 50\% for test. Due to the large sample size, the SVM model is not applicable, and we compare the proposed xNN.enhanced model with the LogR, RF, MLP, and xNN.naive model. 
	
	The original loan purpose variable is categorical with multiple categories. To reduce complexity, we group these various purposes into five categories, including ``Credit Card'', ``Debt Consolidation'', ``Housing'', ``Purchase'' and ``Others''. A summary of the dataset is given in Table~\ref{LC_Description}. This categorical variable is preprocessed using one-hot-encoding, such that each category corresponds to a different bias term in the subnetwork. These bias terms are then added up to form the subnetwork output, which is then linked to the final output, subject to the sparsity constraint.
	
\begin{figure}[!t]
		\centering
		\includegraphics[width=0.4\textwidth]{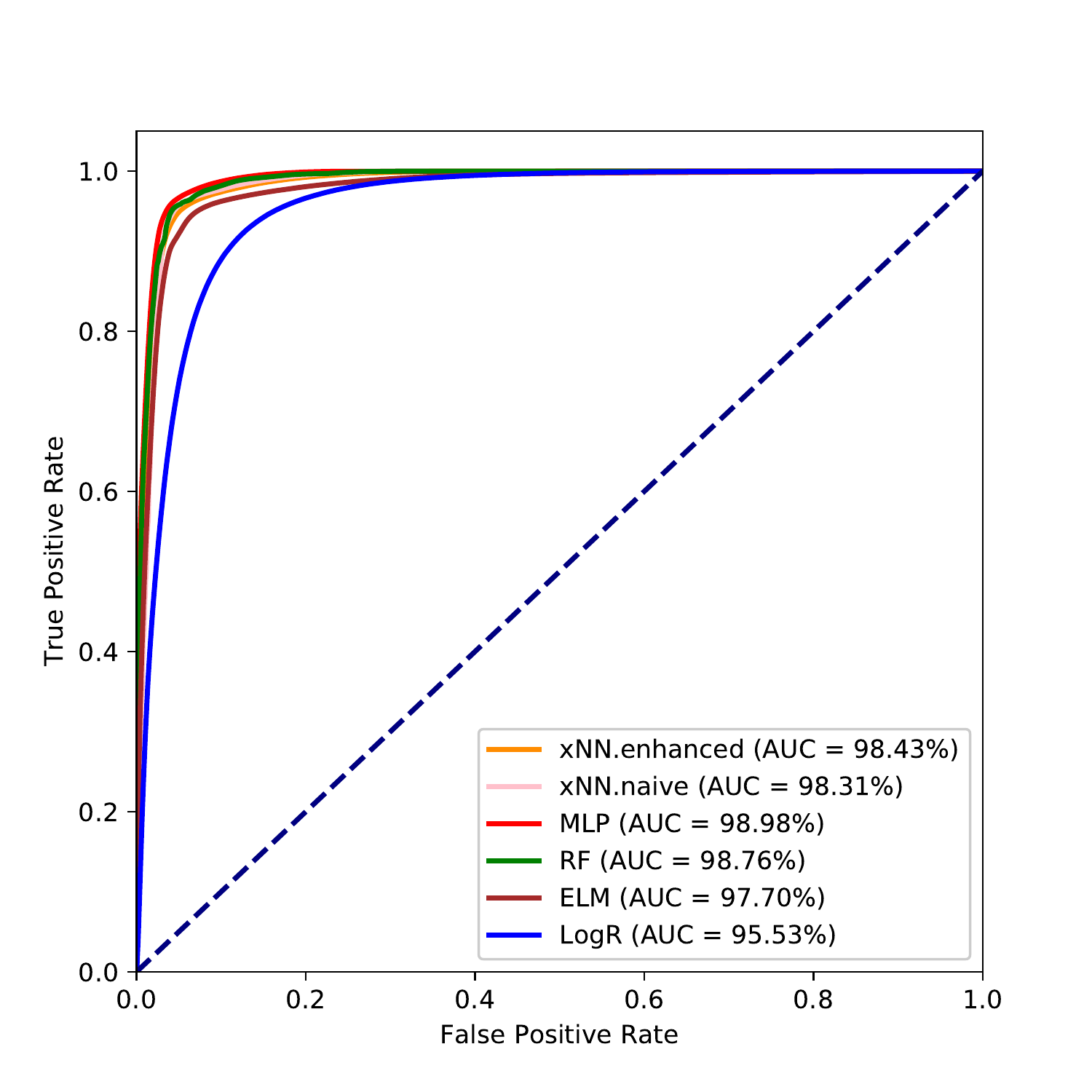}
		\caption{ROC curve comparison for the LendingClub dataset.} \label{LC_OXNN_ROC}
	\end{figure}

	The receiver operating characteristic (ROC) curves of different models are plotted in Fig.~\ref{LC_OXNN_ROC}, together with the area under curve (AUC) scores shown in the bottom right corner. The MLP model performs the best while the LogR model ranks the last. The other models show a relatively close performance. However, despite the high predictive performance, both the black-box models, including the RF, MLP, and ELM, are too complex to interpret; the LogR can be easily interpreted, but it is less accurate. In contrast, the proposed xNN.enhanced model achieves relatively high accuracy, while the estimated model is essentially interpretable. 
	
	\begin{table}[!t]
		\begin{center}
			\caption{Description of the LendingClub Dataset.} \label{LC_Description}
			\begin{tabular}{ccc}
				\hline
				No. &     Variables     &         Range                         \\ \hline
				X1  & Application Date  & Jan. 2015 - Jun. 2018                 \\
				X2  & Amount Requested  &       150 - 40000                    \\
				X3  &    Risk Score     &       300 - 850                       \\
				X4  & Debt Income Ratio &       0\% - 200\%                     \\
				X5  & Employment Length &         0 - 10 \& 11 (more than 10)   \\
				X6  &   Loan Purpose    & \tabincell{c}{``Debt Consolidation'',
					``Credit Card'',  \\ ``Housing'', ``Purchases'', and ``Others''} \\ \hline
			\end{tabular} 
		\end{center}
	\end{table}
	
\begin{figure}[!t]
		\centering
		\subfloat[Marginal plot]{
			\label{GT_LC} 
			\includegraphics[width=0.22\textwidth,height=0.4\textheight]{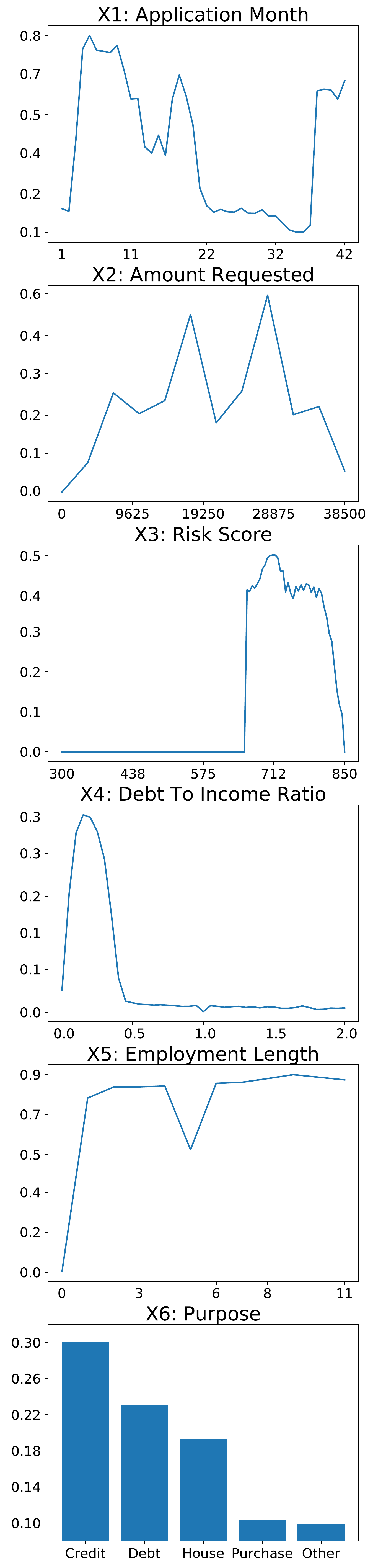}
		}
		\subfloat[xNN.enhanced]{
			\label{LC_SOSxNN} 
			\includegraphics[width=0.4\textwidth, height=0.4\textheight]{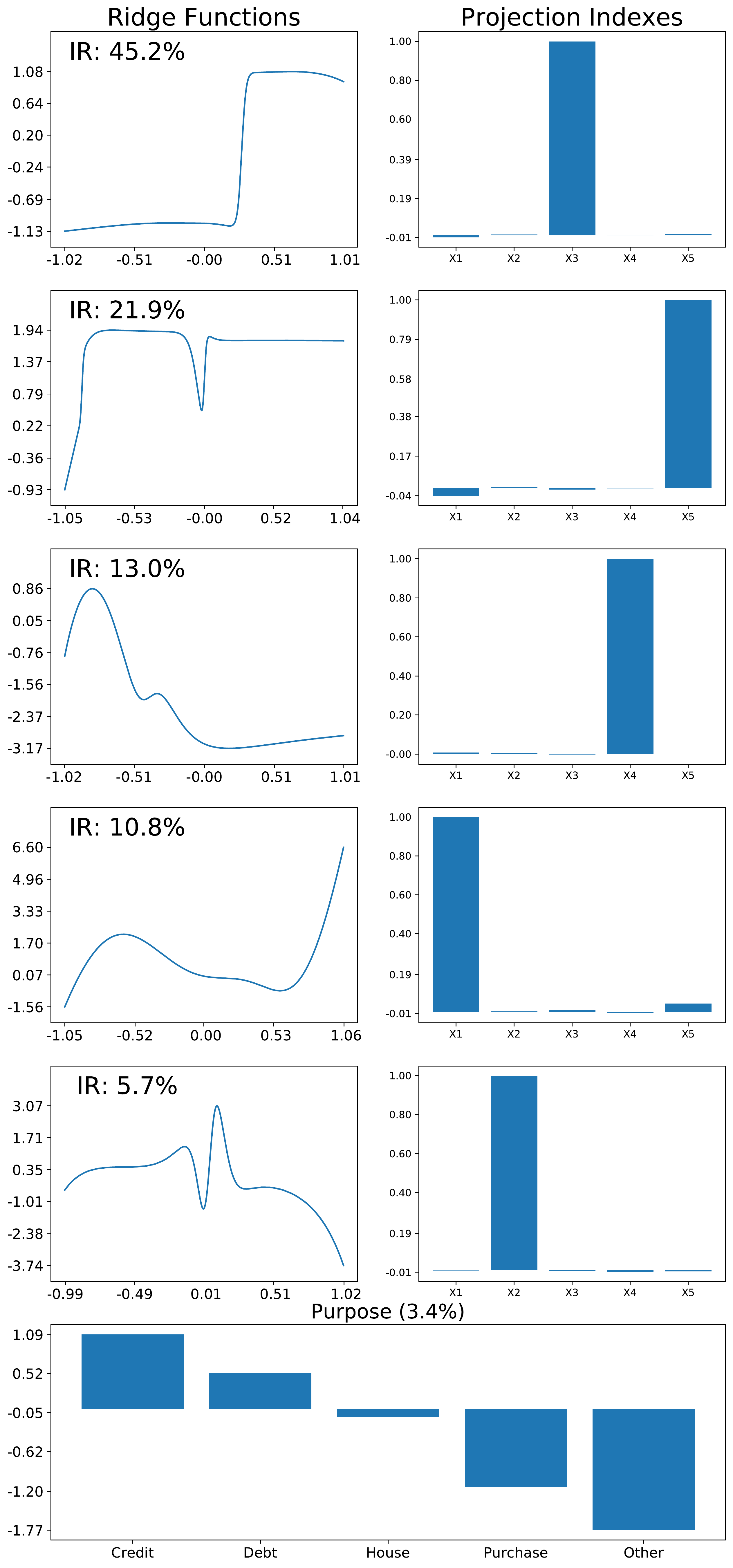}
		}
		\caption{The left sub-figure shows the marginal approved rates with respect to both numerical and categorical variables. The right sub-figure shows the estimated xNN.enhanced model.} \label{LC_OXNN_Vis}
	\end{figure}

	The estimated subnetworks and projection indexes of xNN.enhanced are visualized in Fig.~\ref{LC_OXNN_Vis}. Without any prior knowledge, xNN.enhanced automatically approximates the GAM structure. Compared to marginal approved rates on the left subfigure of Fig.~\ref{LC_OXNN_Vis}, the estimated ridge functions are smooth and can provide a quantitative ranking of different variables. More specifically, we have the following findings. 
	
	First, the risk score (X3) is the most important feature for a loan application, with 45.2\% importance ratio. It is found that the risk scores around 700 are generally preferred. Second, the employment length (X5) accounts for the second important feature (with 21.9\% importance ratio). It will impact applicants who have fewer than two years of working experience and also a slight influence for applicants who have around 5 years of working experience. Third, the model suggests that a moderate debt-income-ratio (X4) is preferable. The application date (X1) is also an influential factor of the application result, as the loan acquisition policy may change over time. Finally, the loan amount (X2) and loan purposes (X6) turn out to be less important, but we can still get hints for successful applications. For instance, the loan applications with purpose of ``Credit card'' and ``Debt consolidation'' are shown to have higher approved rates than the other three categories. 
	
	\subsection{Summary of Results}
	We summarize the above experimental results from the following two perspectives.
	\begin{enumerate}[a)]
		\item The proposed xNN.enhanced model is flexible enough to decompose a complex relationship into several additive components. Compared with the popularly used machine learning algorithms, the proposed model is competitive in the sense of prediction accuracy. The numerical results show that the xNN.enhanced model is competitive regarding prediction accuracy.
		
		\item The proposed xNN.enhanced model is inherently interpretable with additive, sparsity, orthogonality, and smoothness considerations. The estimated projection indexes are sparse and uncorrelated, and the subnetworks are smooth. When the true underlying model is close to the additive assumption, most of the main effects can be nicely captured or well explained; in more complicated settings, the xNN.enhanced model can be an interpretable surrogate, which enhances the interpretability of neural networks for modeling complex relationships.
	\end{enumerate}
	Therefore, the proposed xNN.enhanced method provides an effective approach that balances between predictive accuracy and model interpretability. It is justified that xNN.enhanced is a promising tool for interpretable machine learning. 
	
	\section{Conclusion} \label{Conclusion}
	This paper studies the explainable neural networks subject to interpretability constraints in terms of additivity, sparsity, orthogonality, and smoothness.  First, a complex function is decomposed into sparse additive subnetworks, which is straightforward for model interpretation. Second, the projection indexes are enforced to be mutually orthogonal such that the resulting subnetworks tend to be less confounded with each other. Lastly, each subnetwork-represented ridge function is estimated subject to the smoothness constraint, so that the functional relationship can be better explained. Numerical experiments based on simulation and real-world datasets demonstrate that the proposed xNN model  has competitive prediction performance compared to existing machine learning models. More importantly, such an xNN model is designed to be intrinsically interpretable.
	
	There are some promising topics for future study. First, the xNN architecture can be further improved to take other practical constraints into account. For example, the ridge function can be constrained to be monotonic (either increasing or decreasing), so that the resulting model can fit better the prior experience or domain knowledge. Second, the orthogonality constraint (\ref{ConD}) might be too restrictive to achieve high prediction accuracy, and it is realistic to consider the near-orthogonality as a relaxed constraint. Third, it is also of our interest to investigate the main effects and interaction effects under the xNN framework. It is our plan to study the sensitivity of these effects in the sense of functional analysis of variance.
	
	\section*{Appendix}	
	\subsection*{Proof of Theorem 1} \label{proof_theorem1}

	When Condition (B1) holds, and the projection matrix is orthogonal, then we can directly deduce that Condition (A1) and Condition (A3) hold. According to \cite{yuan2011identifiability}, the sum of the quadratic components is identifiable while the individual quadratic components are not identifiable. Here we are in the position to prove that Condition (B2) is sufficient to identify the individual quadratic ridge functions.
	
	Without loss of generality, assume the first $m$ ridge functions are quadratic, written as $h_{j}(z) = \alpha_{j}z^{2}$, where the non-zero coefficients $\alpha_{1} > \alpha_{2} > \ldots > \alpha_{m}$ for $m\leq k$. Using matrix notation, these quadratic terms can be represented by a $p\times p$ matrix $ \bm{U} = \sum_{j=1}^{m} \alpha_{j}\w_{j}\w_{j}^{T}$. Since $U$ is a real symmetric matrix,  we have that
	$$
	\bm{U} \w_{i} =\sum_{j=1}^{m} \alpha_{j}\w_{j}\w_{j}^{T} \w_{i}
	= \alpha_{i} \w_{i}
	$$
	for $ i = 1, 2,\ldots, m$. Construct a $p \times p$ diagonal matrix $\bm{\Lambda} = \mbox{diag}(\alpha_{1}, \alpha_{2}, \ldots, \alpha_{m}, 0, \ldots, 0)$ and a $p \times p$ orthonormal matrix $\bm{Q} = [\w_{1}, \ldots, \w_{m}, \v_{m+1}, \ldots, \v_{p} ]$ with arbitrary orthogonal complement $\{\v_{m+1}, \ldots, \v_{p}\}$ to $\{\w_{1}, \ldots, \w_{m}\}$. Then, we can write $\bm{U} = \bm{Q \Lambda Q^{T}}$, which corresponds to the eigen-decomposition of $\bm{U}$. The pairs of $(\alpha_{i}, \w_{i})$ are the non-zero eigenvalues and corresponding eigenvectors, respectively. Since $(\alpha_{1}, \alpha_{2}, \ldots, \alpha_{m})$ are all distinct, the corresponding unit eigenvectors are uniquely determined up to sign. Therefore, both the ridge functions and projection indexes of these quadratic components are identifiable.
	
	\section*{Acknowledgment}	
	We thank Vijay Nair and Joel Vaughan from Wells Fargo for insightful discussions. This research project was partially supported by the Big Data Project Fund of The University of Hong Kong from Dr Patrick Poon's donation.

\end{document}